\documentclass[letterpaper]{article} 
\usepackage[submission]{aaai24}  
\usepackage{times}  
\usepackage{helvet}  
\usepackage{courier}  
\usepackage[hyphens]{url}  
\usepackage{graphicx} 
\usepackage{natbib}  
\usepackage{caption} 
\usepackage{algorithm}
\usepackage{algorithmic}

\usepackage{cuted}

\usepackage{amsmath}
\usepackage{amsfonts}
\usepackage{amsthm}
\usepackage{amssymb}
\usepackage{mathtools}

\theoremstyle{plain}
\newtheorem{theorem}{Theorem}

\newtheorem{corollary}[theorem]{Corollary}

\theoremstyle{definition}
\newtheorem{definition}[theorem]{Definition}

\theoremstyle{remark}

\newtheorem*{theorem*}{Theorem}
\newtheorem*{lemma*}{Lemma}
\newtheorem*{definition*}{Definition}
\newtheorem*{corollary*}{Corollary}
\newtheorem*{remark*}{Remark}

\DeclareMathOperator*{\E}{\mathbb{E}}

\newcommand{\s}{\mathcal{S}}

\newcommand{\A}{\mathcal{A}}

\newcommand{\T}{\mathcal{T}}

\usepackage{newfloat}
\usepackage{listings}
\DeclareCaptionStyle{ruled}{labelfont=normalfont,labelsep=colon,strut=off} 
\floatstyle{ruled}
\newfloat{listing}{tb}{lst}{}
\floatname{listing}{Listing}
\author{
    Jacob Adamczyk\textsuperscript{\rm 1}, Stas Tiomkin\textsuperscript{\rm 2}, Rahul V. Kulkarni\textsuperscript{\rm 1}
}
\affiliations{
    \textsuperscript{\rm 1}Department of Physics, University of Massachusetts Boston\\

    \textsuperscript{\rm 2}Department of Computer Engineering, San Jos\'e State University
    
    jacob.adamczyk001@umb.edu, stas.tiomkin@sjsu.edu, rahul.kulkarni@umb.edu

}

\usepackage{color}
\usepackage{multicol}

\usepackage{etoolbox}
\newtoggle{appendix}
\togglefalse{appendix}

\iftoggle{appendix}{
\title{Technical Appendix: \\Leveraging Prior Knowledge in Reinforcement Learning via \\Double-Sided Bounds on the Value Function }
}{
\title{Leveraging Prior Knowledge in Reinforcement Learning via \\Double-Sided Bounds on the Value Function }
}

\begin{document}

\def\CaoThm{Theorem~1 }
\def\CaoThmNospace{Theorem~1}
\def\HaarLem{Lemma 1 }
\def\zeroshot{zero-shot}

\newcommand\av[1]{\left\lvert#1\right\rvert}
\newcommand{\fr}{f\left(\{r_j (s,a)\}\right)}
\newcommand{\fQ}{f\left(\{Q_j^* (s,a)\}\right)}
\newcommand{\G}{\mathcal{G}}

\newcommand{\fQspap}{f\left(\{Q_j^* (s',a')\}\right)}
\theoremstyle{plain}
\newtheorem{innercustomgeneric}{\customgenericname}
\providecommand{\customgenericname}{}
\newcommand{\newcustomtheorem}[2]{%
  \newenvironment{#1}[1]
  {%
   \renewcommand\customgenericname{#2}%
   \renewcommand\theinnercustomgeneric{##1}%
   \innercustomgeneric
  }
  {\endinnercustomgeneric}
}

\newcustomtheorem{customthm}{Theorem}
\newcustomtheorem{customlemma}{Lemma}
\newcustomtheorem{customcor}{Corollary}

\def\AAAITheorem#1{
    \begin{customthm}{5.1}[\cite{Adamczyk_AAAI}]
        Let a set of primitive tasks $\{\T_j\}_{j=1}^{M}$
        with corresponding optimal value functions $\{Q_j^*\}$ be given.
        Denote $\widetilde{Q}^*$ as the optimal action-value function for the task $\widetilde{\T}$ with reward function $\widetilde{r}(s,a)$. Define $K^*$ as the optimal soft action-value function for a task with reward function $\kappa$ and prior policy $\pi_f$ with the following definitions:
        \begin{align*}\label{eq:rwd for composition}
            \kappa(s,a) = \widetilde{r}(s,a) + \gamma \E_{s' \sim{} p} V_f(s') - \fQ,
        \end{align*}
        \begin{equation*}
            V_f(s) = \frac{1}{\beta}\log\E_{a\sim{} \pi_0}\exp{\beta \fQ },
        \end{equation*}
        \begin{equation}
            \pi_f(a\vert s) = \pi_0(a\vert s) \frac{e^{\beta \fQ }}{e^{\beta V_f(s)}}.
        \end{equation}
        Then, the optimal value function $\widetilde{Q}^*$ and the \zeroshot\ solution $f(\{Q_j^*\})$ are related by:
        \begin{equation}
            \widetilde{Q}^*(s,a) = \fQ + K^*(s,a)
            \label{eq:qtilda=f(q)+K}
        \end{equation}
        \label{thm:aaai#1}
    \end{customthm}
}

\def\StdRLRwdChange#1{
    \begin{customthm}{6.1a}[\citeauthor{ng_shaping}]
        Let a (standard RL) primitive task $\T$ with reward function $r$ be given, with the optimal value function $V^*(s)$.
        Consider another (standard RL) task, $\widetilde{\T}$ with reward function $\widetilde{r}$, with an unknown optimal action-value function, $\widetilde{Q}^*$.
        Define $\kappa(s,a) \doteq \widetilde{r}(s,a) + \gamma \E_{s'} V^*(s') - V^*(s)$.\\
        Denote the optimal action-value function $K^*$ as the solution of the following Bellman optimality equation
        \begin{equation}\label{eq:K_backup1#1}
            K^{*}(s,a) = \kappa(s,a) + \gamma \E_{s' \sim{} p} \max_{a'} K^*(s',a')
        \end{equation}
        Then, \begin{equation}\label{eq:Q=Q+K rwd_change#1}
            \widetilde{Q}^*(s,a) = V^*(s) + K^*(s,a)
        \end{equation}
        \label{thm:rwd_change_stdRL#1}
    \end{customthm}
}

\def\StdRLComp#1{
    \begin{customthm}{4.8}
        Given a set of primitive tasks $\{\T_j\}$
        with corresponding optimal value functions $\{Q_j^*\}$, denote
        $\widetilde{Q}^*$ as the optimal value function for the composition
        of $\{\T_j\}$ under the composition function $f: \mathbb{R}^M \to \mathbb{R}$.

        Define $K^*$ as the optimal value function for a task with reward function $\kappa$ defined by:
        \begin{align*}\label{eq:rwd for composition}
            \kappa(s,a) = f(\{r_j(s,a)\}) + \gamma \E_{s'} V_f(s') - V_f(s)
        \end{align*}
        \begin{equation*}
            V_f(s) = \max_a \fQ
        \end{equation*}

        Then, the optimal value functions $\widetilde{Q}^*$ and $K^*$ are related by:
        \begin{equation}
            \widetilde{Q}^*(s,a) = V_f(s) + K^*(s,a)
            \label{eq:std RL qtilda=f(q)+K#1}
        \end{equation}
        \label{thm:std_aaai_#1}
    \end{customthm}
}

\def\ClippingCorollaryExact#1{
    \begin{customcor}{5.2}[Zero-Shot Bounds]
        Let a bounded composition function $f: \mathbb{R}^M \to \mathbb{R}$ be given. The optimal value function
        for the composite task, $\widetilde{Q}^*(s,a)$, is upper- and lower-bounded at the state-action level:
        \begin{align}
            \widetilde{Q}^*(s,a) & \leq \fr + \gamma \left(\E_{s'}V_f(s') + \frac{\max \kappa}{1-\gamma} \right) \label{eq:fq_K_bound1#1} \\
            \widetilde{Q}^*(s,a) & \geq \fr + \gamma \left(\E_{s'}V_f(s') + \frac{\min\kappa}{1-\gamma} \right) \label{eq:fq_K_bound2#1} 
        \end{align}
        \label{cor:clipping_exact#1}
    \end{customcor}
}

\def\SuboptimalityTheoremComp#1{
    \begin{customthm}{5.3}
        Let $\pi_f$ denote the \zeroshot\ policy obtained from the primitive value functions $\{Q^*_j\}$:
        \begin{equation}
            \pi_f(a\vert s) = \pi_0(a\vert s) \frac{e^{\beta \fQ }}{e^{\beta V_f(s)}}.
        \end{equation}

        Consider the soft value of the policy $\pi_f$ obtained from deploying it in the composite task:
        $\widetilde{Q}^{\pi_f}(s,a)$, and denote $\widetilde{V}^{\pi_f}=1/\beta \log \E_{a' \sim{} \pi_0} \exp \beta \fQ$.
        Let $\hat{D}^*$ be the optimal soft value function for the task with prior policy $\pi_f$ and reward function
        \begin{equation}
            \hat{d}(s,a) = \fr + \gamma \E_{s' \sim{} p} \widetilde{V}^{\pi_f}(s') - \widetilde{Q}^{\pi_f}(s,a)
        \end{equation}
        Then, the value functions satisfy:
        \begin{equation}
            \widetilde{Q}^{\pi_f}(s,a) = \widetilde{Q}^*(s,a) - \hat{D}^*(s,a)
        \end{equation}
        \label{thm:pi_f#1}\end{customthm}

}
\def\SuboptBound#1{
    \begin{customcor}{4.2}[Suboptimality Bounds]
        Let policy $\pi(a|s)$ be given with soft value $Q^{\pi}(s,a)$. The rate of the suboptimality gap, $Q^*(s,a) - Q^\pi(s,a)$, is then bounded between
        \begin{equation}
        \inf_{(s,a)} d(s,a) \leq \frac{Q^*(s,a) - Q^\pi(s,a)}{H} \leq \sup_{(s,a)} d(s,a)
        \end{equation}

        where $d(s,a) \doteq r(s,a) + \gamma \E_{s'} V^{\pi}(s') - Q^{\pi}(s,a)$, $V^{\pi}(s) \doteq \log \E_{a} \exp \beta Q^{\pi}(s,a)$ is the soft state-value function, and $H = (1-\gamma)^{-1}$ is the effective time horizon.
        \label{cor:regret_bounds#1}
    \end{customcor}
}

\def\SuboptimalityTheoremHaarnoja#1{

    \begin{corollary}
        With the definitions of Theorem~\ref{thm:pi_f}, the value of $\pi_f$ satisfies
        \begin{equation}
            \widetilde{Q}^{\pi_f}(s,a) \geq \widetilde{Q}^*(s,a) - D^*(s,a)
        \end{equation}
        where $D^*(s,a)$ is the standard RL ($\beta \to \infty$ limit) optimal value function
        corresponding to the reward function $\hat{d}$ of Theorem \ref{thm:pi_f}.
    \end{corollary}

}

\def\OnTheFlyBoundComposite#1{
    \begin{customthm}{5.5}
        Let $\pi_k$ be the policy obtained at step $k$ during training of the composite value function, $\widetilde{Q}^*(s,a)$.

        Consider the corresponding soft-value of the policy $\pi_k$ obtained by deploying it in the composite task's environment:
        $\widetilde{Q}^{\pi_k}$, and denote $\widetilde{V}^{\pi_k}~=~1/\beta \log \E_{a \sim \pi_0} \exp \beta \widetilde{Q}^{\pi_k}(s,a)$.
        Define the reward function
        \begin{equation}
            \hat{d}_k(s,a) = \fr + \gamma \E_{s' \sim{} p} \widetilde{V}^{\pi_k}(s') - \widetilde{Q}^{\pi_k}(s,a)
        \end{equation}
        Then the optimal value function for the composite task, $\widetilde{Q}^*$ is bounded by:
        \begin{align}
            \widetilde{Q}^*(s,a) & \leq \fr + \gamma \left(\E_{s' \sim{} p} \widetilde{V}^{\pi_k}(s') + \frac{\max{\hat{d}_k}}{1-\gamma}\right)  \\
            \widetilde{Q}^*(s,a) & \geq \fr + \gamma \left(\E_{s' \sim{} p} \widetilde{V}^{\pi_k}(s') + \frac{\min{\hat{d}_k}}{1-\gamma}\right)
            \label{eq:pi_k#1}
        \end{align}
        \label{thm:pi_k#1}
    \end{customthm}

}

\def\OnTheFlyBound#1{
    \begin{customthm}{4.1}[]
        Consider an entropy-regularized MDP $\langle \s, \A, p, r, \gamma, \beta \rangle$ with (unknown) optimal value function $Q^*(s,a)$.
        Let an estimate for the value function $Q(s,a)$ be given.
        Denote $V(s)~\doteq~1/\beta \log \E_{a \sim \pi_0} \exp \beta Q(s,a)$.

        The optimal value function $Q^*(s,a)$ is then bounded by:
        \begin{subequations}
        \begin{align}
            Q^*(s,a) &\geq r(s,a) + \gamma \left(\E_{s' \sim{} p} V(s') + \frac{\inf \Delta}{1-\gamma}\right) \label{eq:gen_double_sided_boundsA#1}\\
            Q^*(s,a) &\leq r(s,a) + \gamma \left(\E_{s' \sim{} p} V(s') + \frac{\sup\Delta}{1-\gamma}\right) \label{eq:gen_double_sided_boundsB#1}
        \end{align}
        \end{subequations}
        where         
        \begin{equation*}
            \Delta(s,a) \doteq r(s,a) + \gamma \E_{s' \sim{} p} V(s') - Q(s,a).
        \end{equation*}
    \label{thm:gen_double_sided_bounds#1}
    \end{customthm}
    In Eq. \eqref{eq:gen_double_sided_boundsA#1} and \eqref{eq:gen_double_sided_boundsB#1}, the $\inf$ and $\sup$ are taken over the continuous state-action space $\s \times \A$.
}

\def\OnTheFlyBoundStd#1{
    \begin{customthm}{4.9}
        Consider a (standard RL) task with reward function $r(s,a)$ and (unknown) optimal value function $Q^*(s,a)$.
        Let an estimate for the state value function be given as $V(s)$.

        The optimal value function $Q^*(s,a)$ is then bounded by:
        
        \begin{align}
            Q^*(s,a) &\geq r(s,a) + \gamma \left(\E_{s' \sim{} p} V(s') + \frac{\inf \Delta}{1-\gamma}\right) \\
            Q^*(s,a) &\leq r(s,a) + \gamma \left(\E_{s' \sim{} p} V(s') + \frac{\sup\Delta}{1-\gamma}\right) \label{eq:std_gen_double_sided_bounds#1}
        \end{align}
        where         
        \begin{equation*}
            \Delta(s,a) \doteq r(s,a) + \gamma \E_{s' \sim{} p} V(s') - V(s).
        \end{equation*}
    \label{thm:std_gen_double_sided_bounds}
    \end{customthm}
    In Eq. \eqref{eq:std_gen_double_sided_bounds#1}, the $\inf$ and $\sup$ are taken over the continuous state-action space $\s \times \A$.
}

\def\TightOnTheFlyBound#1{
    \begin{customlemma}{4.1a}[]
        Consider an entropy-regularized MDP $\langle \s, \A, p, r, \gamma, \beta \rangle$ with (unknown) optimal value function $Q^*(s,a)$.
        Let an estimate for the value function $Q(s,a)$ be given.
        Denote $V(s)~\doteq~1/\beta \log \E_{a \sim \pi_0} \exp \beta Q(s,a)$. Suppose there exists an ``identity'' action $a_\emptyset(s) \in \A$ for each state, which deterministically transitions the agent to the same state: $p(s'|s,a_\emptyset(s) ) = \delta(s'-s)$ for all $s \in \s$.

        Then the lower bound on the optimal value function $Q^*(s,a)$ can be improved:
        
        \begin{equation}
            Q^*(s,a) \geq r(s,a) + \gamma \left( V(s') + \frac{1}{1-\gamma}\Delta(s', a_\emptyset) \right)
            \label{eq:tighter_double_sided#1}
        \end{equation}
    \end{customlemma}
}

\def\PolicyImprovementEqn#1{
    \begin{customthm}{5.6}
        Given a policy $\pi_k$ in entropy-regularized RL with soft value $Q^{\pi_k}(s,a)$, then the improved policy $\pi_{k+1} \propto \exp \beta Q^{\pi_k}(s,a)$ has a soft value, $Q^{\pi_{k+1}}(s,a)$ given by:
        \begin{equation}
            Q^{\pi_{k+1}}(s,a) = Q^{\pi_{k}}(s,a) + q^{\pi_{k+1}}(s,a)
        \end{equation}
        where $q^{\pi_{k+1}}(s,a)$ is the value for policy $\pi_{k+1}$ in standard RL for a task with reward function $r(s,a) = r(s,a) + \gamma \E_{s'}V^{\pi_k}(s') - Q^{\pi_k}(s,a) $ 
    \label{thm:PI_stdPE}
\end{customthm}
}

\def\BoundingContinuousExtrema#1{
    \begin{customlemma}{4.4}
        Let $\s \times \A$ be a bounded metric space with diameter $D$, 
        and let $r: \s \times \A \to \mathbb{R}$ be $L_r$-Lipschitz (w.r.t. the same metric). 
        Then the global extrema of $r(s,a)$ on $\s \times \A$ are bounded as follows:
        \begin{align*}
            \sup_{s\in \s,a \in \A} r(s,a) &\leq \min_{(s,a) \in \mathcal{D}}  r(s,a) + L_r D  \\ 
            \inf_{s\in \s,a \in \A} r(s,a) & \geq \max_{(s,a) \in \mathcal{D}} r(s,a) - L_r D
        \end{align*}
        where $\mathcal{D}$ is the dataset of $(s,a)$ tuples available for querying the magnitude of $r$ (e.g. the current batch or buffer).
        \label{lem:extrema#1}
    \end{customlemma}
}

\def\VfuncBound#1{
\begin{customthm}{4.5}

Let an entropy-regularized MDP be given with an $L_Q$-Lipschitz value function $\bar{Q}^\pi$. Using a Gaussian parameterization for the associated policy $\pi(\cdot|s)=\mathcal{N}(\mu(s),\sigma(s))$, suppose that $\bar{Q}^\pi$ is an $\varepsilon$-optimal approximation of the policy's true value, $Q^\pi$.

By estimating the state-value function as:
\begin{equation}
    \bar{V}^\pi(s) = \bar{Q}^\pi(s,\mu) - \frac{1}{\beta} \E_{a\sim{} \pi} \log\frac{\pi(a|s)}{\pi_0(a|s)},
    \label{eq:approx_V#1}
\end{equation}

the error in using such an approximation is upper bounded:
\begin{equation*}
    |\bar{V}^\pi(s) - V^\pi(s)| \leq \sqrt{\frac{2 }{\pi}}L_Q \sigma(s) e^{-\mu(s)^2/2\sigma(s)^2}  + \varepsilon
\end{equation*}
In the case that the function $Q$ used is an optimal value function for an $(L_r, L_p)$-Lipschitz task, with a policy whose variance is lower bounded $\sigma(s) \geq \sigma_\text{min}$ and $\gamma L_p (1+L_\mathcal{N})<1$, where $L_\mathcal{N} =\sigma_{\text{min}}^{-2}(2\pi e)^{-1/2}$ is the Lipschitz constant of the Gaussian distribution, then the Lipschitz constant for $Q$ can be computed as:
\begin{equation}
    L_Q=\frac{L_r + \gamma L_p (\beta \sigma_{\min})^{-1}}{1-\gamma L_p (1+L_\mathcal{N})}.
    \label{eq:L_Q_entreg#1}
\end{equation}


\label{thm:vfunc_bound#1}
\end{customthm}
}
\def\Bogoliubov#1{
    \begin{customthm}{5.3}
        \begin{equation}
            \widetilde{Q}^*(s,a) \geq \E_\pi \sum_{t=0}^{\infty} \gamma^t \left( \widetilde{r}(s_t, a_t) + \frac{1}{\beta} \textrm{KL}(\pi|\pi_0) \right)
        \end{equation}
        \label{lem:theorem}
    \end{customthm}

}

\def\ClippedBOperatorThm#1{
\begin{customthm}{4.3}
Let the functions $L(s,a), U(s,a)$ be lower and upper bounds on the optimal value function: $L(s,a)~\leq~Q^*(s,a)~\leq U(s,a)$ for all $s \in \s$ and $a\in\A$. The clipped Bellman operator, $\mathcal{B}_{C}Q(s,a)~:=~\max_{s,a} \left( \min_{s,a} \left(\mathcal{B}Q(s,a), U(s,a) \right), L(s,a) \right)$ converges to the optimal value function $Q^*(s,a)~=~\mathcal{B}^\infty Q(s,a)$. 
\end{customthm}
}

\def\vanNiekerkExtension#1{
\begin{customthm}{5.1}
    Consider $m$ solved tasks in the entropy-regularized setting, with reward functions $\{r_1,\dotsc,r_m\}$ varying only on the set of absorbing states. Assume all tasks are given with the same deterministic dynamics. Given a set  of non-negative weights $w_j$, consider a new task with the same reward function for the interior (i.e. non-absorbing) states and with reward function for the absorbing states given by
    \begin{equation}
        \widetilde{r}(s,a) = \tau \log \sum_{j=1}^m w_j e^{r_j(s,a) / \tau}.
    \end{equation}
    Then, the optimal value function for such a task is given by:
    \begin{equation}
    \widetilde{Q}(s,a) = \tau \log \sum_{j=1}^m w_j e^{Q_j(s,a) / \tau}.
    \end{equation}
    \label{thm:niek}
\end{customthm}
}

\def\FullyPropagated#1{
\begin{customthm}{4.6}
Let the $L_Q$-Lipschitz value function $Q^\pi$ and corresponding Gaussian policy $\pi(\cdot|s) = \mathcal{N}(\mu(s), \sigma(s))$ be given, where $Q^\pi$ is an $\varepsilon$-optimal estimate of the true policy's value function.
For an $(L_r, L_p)$-Lipschitz task with (unknown) optimal value function $Q^*$, let $\bar{V}^\pi$ be the one-point estimate of the (known) value function $Q^\pi$, and denote $\bar{\Delta}(s,a) = r(s,a) + \gamma \E_{s' \sim{} p} \bar{V}^\pi(s') - Q^\pi(s,a)$. Then:
\begin{align*}
    &Q^*(s,a) \leq r(s,a) + \gamma \E_{s' \sim{} p}\left[\bar{V}^\pi(s') + A(s')\right] \\     
    &\hspace{1em}+\frac{\gamma}{1-\gamma}\left(\min_{(s,a) \in \mathcal{D}} \left(\bar{\Delta}(s,a) + \gamma \E_{s'\sim{} p} A(s') \right) + L_\Delta D  \right) \\
    &Q^*(s,a) \geq r(s,a) + \gamma \E_{s' \sim{} p}\left[\bar{V}^\pi(s') - A(s')\right] \\     
    &\hspace{1em}+\frac{\gamma}{1-\gamma}\left(\max_{(s,a) \in \mathcal{D}} \left(\bar{\Delta}(s,a) - \gamma \E_{s'\sim{} p}A(s') \right) - L_\Delta D  \right) \\%
\end{align*}
where we let $A(s)= \sqrt{\frac{2}{\pi}}L_Q\sigma(s) e^{-\mu(s)/2\sigma(s)^2}+\varepsilon$ and 
$L_\Delta=\max \left\{L_r, L_Q, \gamma L_p \left(L_Q(1+L_\mathcal{N}) + (\beta \sigma_{\text{min}})^{-1}\right)\right\}$ and $D$ 
denotes the diameter of the state-action space.
\label{thm:fully_propagated#1}
\end{customthm}
}

\maketitle
\iftoggle{appendix}{%
\begin{abstract}
In this Appendix, we (1) provide further technical details on the experiments used to demonstrate the results and (2) provide proofs for the theoretical statements in the main text.
\end{abstract}

}{
\begin{abstract}
    An agent’s ability to leverage past experience is critical for efficiently solving new tasks. Approximate solutions for new tasks can be obtained from previously derived value functions, as demonstrated by research on transfer learning, curriculum learning, and compositionality. However, prior work has primarily focused on using value functions to obtain zero-shot approximations for solutions to a new task. In this work, we show how an arbitrary approximation for the value function can be used to derive double-sided bounds on the optimal value function of interest. We further extend the framework with error analysis for continuous state and action spaces. The derived results lead to new approaches for clipping during training which we validate numerically in simple domains.
\end{abstract}
}

\iftoggle{appendix}{
}
{
\section{Introduction}
The field of reinforcement learning (RL) has seen impressive successes  \cite{tokamak,gochess, dsilver1, dsilver2} in recent years due to the development of novel algorithms in combination with deep learning architectures. However, for complex tasks, the amount of training time required for learning an optimal solution from scratch can be prohibitively large and thus presents a significant obstacle to further development. To address this challenge, approaches that leverage prior knowledge to efficiently calculate policies for new tasks are needed. While policies generated from prior solutions may not be the \textit{optimal} policies for the new tasks, they can serve as useful approximations that reduce training time. Correspondingly, there is a need to develop approaches that further leverage the use of approximations based on prior knowledge to address the problem of solving new tasks.

Previous work has focused on addressing this problem using different approaches such as transfer learning, curriculum learning, and compositionality. In particular,
we consider value-based RL approaches, wherein the agent's goal is to learn the expected value of every state and action pair. Given this value function, $Q(s,a)$, the agent can act optimally by choosing actions which maximize its expected future returns.
In many instances, the agent has an estimate for the value function before training begins. For example, in the case of curriculum learning, the agent has the $Q$-values for previously learned (progressively more challenging) tasks. In the case of compositional or hierarchical RL, the agent can combine knowledge by applying a function on subtasks' $Q$-values. When using an exploratory skill-acquisition approach such as DIAYN \cite{eysenbach2018diversity} or CSD \cite{park2023controllabilityaware}, the agent obtains $Q$-values for a diverse set of skills.
Even in cases where an initial estimate is not explicitly provided, the agent can provide itself an estimate by using Q-values that were obtained during the ongoing learning phase (bootstrapping).

An underlying question in these scenarios is the following: How can the agent use the known value function estimate(s) for solving a new target task? Does the estimate only serve as a zero-shot approximation or is there additional useful information that can be extracted from it?

In the work of \cite{Adamczyk_AAAI}, the authors show that there exists a method of ``closing the gap'' between any estimate ($Q^*(s,a)$) and any target ($\widetilde{Q}^*(s,a)$) task (with an accessible reward function) in entropy-regularized RL.
This statement is facilitated by the work of \cite{cao2021identifiability} 
which can be used to show that any estimate can be viewed as an optimal value function corresponding to a suitably defined reward function. 
Here, we show that since the gap between the target and estimated value functions: $\widetilde{Q}^*(s,a)-Q^*(s,a)=K^*(s,a)$ is itself an optimal value function, it can be bounded. As a consequence, instead of providing only a zero-shot approximation or a warmstart for training the target task, we show that the estimates available to the agent also provide a double-sided bound on the optimal $Q$-values being learned. 

A schematic illustration of our approach is provided in Fig.~\ref{fig:schematic}. 
Starting with an estimate of the optimal value function and samples of the reward function, we derive double-sided bounds on the true optimal value function.
We find that applying these bounds during training improves the agent's training performance and allows an additional method for monitoring convergence. 
We provide further theoretical analysis on continuous state-action 
spaces, relevant for the function approximator (FA) setting in Deep RL.

\textbf{Main contributions}
\newline
The main contributions of our work, applicable to both standard and entropy-regularized RL, are:
\begin{enumerate}
    
    \item Development of a general framework for bounding optimal value functions based on prior knowledge.
    \item Extension of derived results to include theoretical error analysis in continuous state-action spaces. 
    \item Demonstration of value-based clipping methods as practical applications of the derived theoretical results.
\end{enumerate}

\begin{figure}
    \centering
    \includegraphics[width=0.47\textwidth]{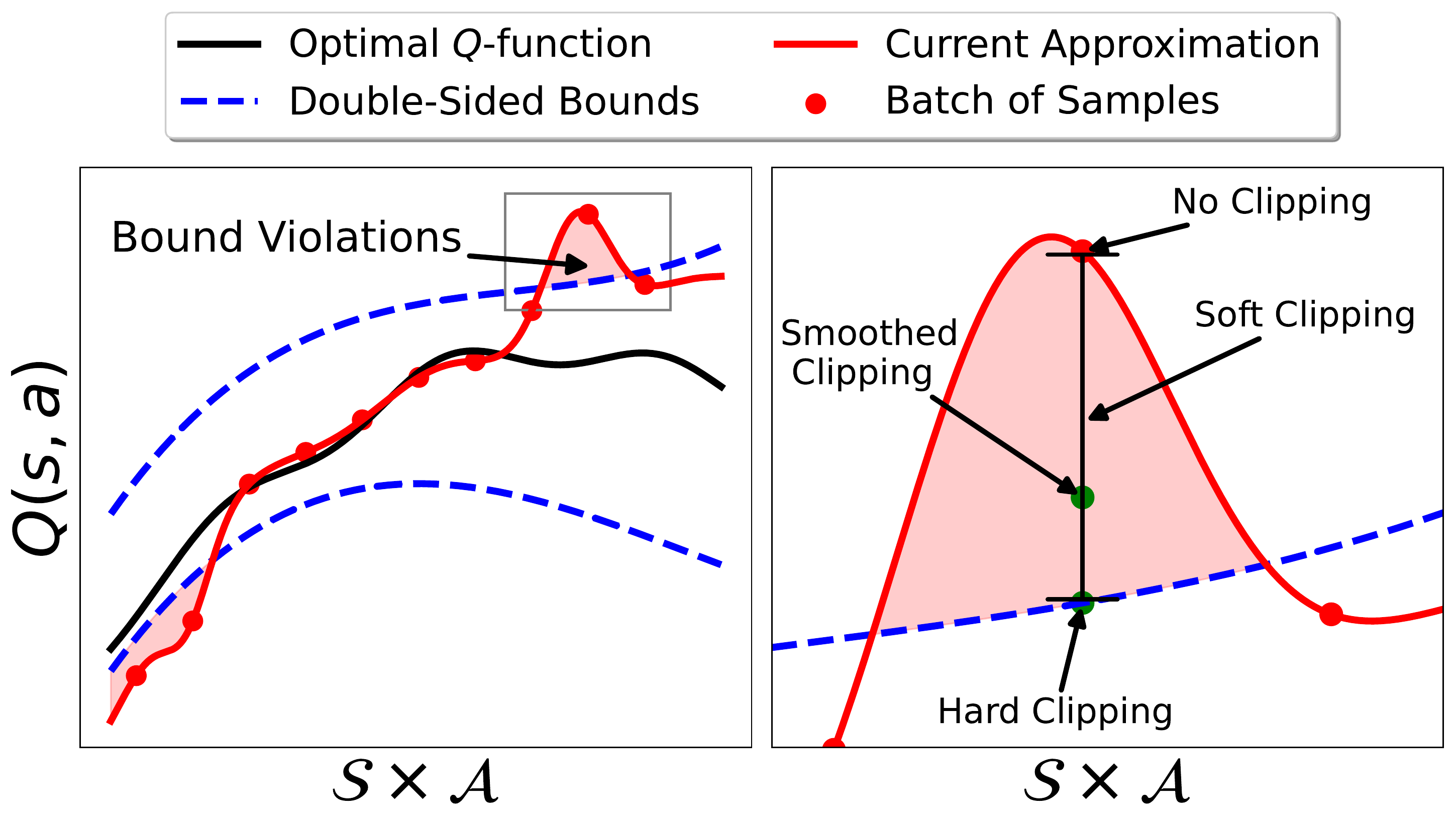}
    \caption{Schematic illustration of the main contribution of this work. Given any approximation (red curve) to the optimal value function of interest (black curve), we derive double-sided bounds (blue curves) that lead to clipping approaches during training. Based solely on the current approximation for $Q(s,a)$ (red curve), we derive double-sided bounds on the \textit{unknown} optimal value function $Q^*(s,a)$ (black curve). In the right panel, we show the different clipping methods, which are described further in the ``Experimental Validation'' section. In ``Hard Clipping'', the target is replaced with the exceeded bound; in ``Soft Clipping'', an additional loss term is appended to the Bellman loss, proportional to the magnitude of the bound violation; in ``Smoothed Clipping'', the target update is replaced with a weighted average of the original value and the exceeded bound.}
    \label{fig:schematic}
\end{figure}

There are multiple applications that arise from the derivation of such double-sided bounds. The bounds (1) allow confinement of FA training to a limited output range, (2) provide a mechanism to choose the ``best'' skill from a pre-trained set of skills and (3) establish a framework that provides insights into and extends previous results on exact compositions of value functions.



\section{Preliminaries}
For the theoretical setup, we consider initially the case of finite, discrete state and action spaces, and we will subsequently extend our analysis to continuous spaces. In this setting, the reinforcement learning (RL) problem is modeled by a Markov Decision Process (MDP) represented as a tuple $\langle \s,\A,p,r,\gamma \rangle$ where $\s$ is the set of available states; $\A$ is the set of possible actions; $p: \s \times \A \to \s$ is the transition function (dynamics); $r: \s \times \A \to \mathbb{R}$ is a (bounded) reward function which associates a reward (or cost) with each state-action pair; and $\gamma \in (0,1)$ is a discount factor which discounts future rewards and assures convergence of the total reward for an infinitely long trajectory.

The objective in standard (un-regularized) RL is to find an optimal policy that maximizes expected rewards collected by the agent, i.e.
\begin{equation}\label{eq:std_rl_obj}
    \pi^* = \arg\max_{\pi} \mathbb{E}
    \left[ \sum_{t=0}^{\infty} \gamma^{t} r(s_t,a_t) \right].
\end{equation}

An important generalization is entropy-regularized RL \cite{ZiebartThesis}, which augments the un-regularized RL objective
(Eq. \eqref{eq:std_rl_obj}) by including an entropic regularization term which penalizes control over a pre-specified
reference policy:
\begin{equation*}
    \pi^* = \arg\max_{\pi}
    \mathbb{E}
    \left[ \sum_{t=0}^{\infty} \gamma^{t} \left( r_t - \frac{1}{\beta}
        \log\left(\frac{\pi(a_t|s_t)}{\pi_0(a_t|s_t)} \right) \right) \right]
\end{equation*}
where $\pi_0(a|s)$ is the fixed prior policy.
The additional control cost discourages the agent from choosing policies that deviate too much from this prior policy.
Importantly, entropy-regularized MDPs lead to stochastic optimal policies that are provably robust to perturbations of rewards and dynamics \cite{eysenbach}; making them a more suitable approach to real-world problems.

The solution to the RL problem is defined by its \textit{optimal} action-value function ($Q^*(s,a)$) from which one can derive the aforementioned \textit{optimal} policy $\pi^*(a|s)$. For both un-regularized and entropy-regularized RL,
the optimal value function can be obtained by iterating a recursive Bellman equation. In un-regularized RL, the Bellman optimality equation is given by \cite{suttonBook}:
\begin{equation}
    Q^*(s,a) = r(s,a) + \gamma \mathbb{E}_{s' \sim{} p(\cdot|s,a)} \max_{a'} \left( Q^*(s',a') \right).
    \label{eq:bellman}
\end{equation}

The entropy term in the objective function of entropy-regularized RL modifies the previous optimality equation in the following way \cite{ZiebartThesis, Haarnoja_SAC}:
\begin{equation}
    Q^*(s,a) = r(s,a) + \frac{\gamma}{\beta} \mathbb{E}_{s' \sim{} p
    }\log \mathbb{E}_{a' \sim{} \pi_0}
    e^{ \beta Q^*(s',a') }.
    \label{eq:soft_bellman}
\end{equation}

The regularization parameter $\beta$ can be interpreted as being analogous to an inverse temperature parameter, its value is used to control the degree of stochasticity in the optimal policy. In the entropy-regularized setting, $Q^*$ is referred to as the optimal ``soft'' action-value function. For brevity, we will hereon refer to $Q^*$ simply as the value function.

\section{Prior Work}

The importance of double-sided bounds on value functions has been explored in prior work. 
In this section we review a set of the most relevant prior works \cite{Nemecek,constrainedGPI,Haarnoja2018,Adamczyk_UAI,Todorov,vanNiekerk,boolean,lee2021sunrise}. We contrast the existing 
works with regard to the following features:
i) the assumption about composition and/or transformation of known solutions in the derivation of  bounds,
ii) the requirement for additional samples needed to derive bounds,
iii) the generality and applicability of bounds to un-regularized RL and entropy-regularized RL, and to deterministic and stochastic dynamics, 
iv) double or single-sided bounds.

In \cite{Nemecek}, the authors have derived double-sided bounds on the state value function $V(s)$ by the \textit{positive conical combination} of subtask rewards. The method in \cite{Nemecek} requires additional samples for first learning the {\it successor features} before then deriving the double-sided bounds for a downstream task. The applicability of \cite{Nemecek} is limited to un-regularized RL.  

The aforementioned work was subsequently extended by \cite{constrainedGPI}, where, in the same GPI setting, they present double-sided bounds on $Q$-values for linear combinations of subtask reward functions. They introduce the notion of ``soft clipping'' which we adapt to our setting (details in the ``Experimental Validation'' section), but it was not demonstrated in practice. Similarly to \cite{Nemecek}, the method in \cite{constrainedGPI} requires firstly to learn {\it the successor features}, and it is limited to un-regularized RL only. 

The previous two works were focused on the standard (un-regularized) reinforcement learning setting. However, the double-sided bounds presented in \cite{Haarnoja2018}'s Lemma 1 are derived for the MaxEnt setting, for the case of convex reward combinations. It is worth noting that the lower bound in this case must be learned (the $C$ function). Extending these results to other more general classes of functional composition, \cite{Adamczyk_UAI} provides double-sided bounds for both entropy-regularized and un-regularized RL. However, one side of the bound in all cases must be learned as well.

Finally, multiple prior works have focused on specific examples of compositionality  for which exact results can be obtained for the optimal value function.
These results typically involve multiple limiting assumptions on the structure of rewards functions, nature of transition dynamics and specific forms for the composition function. \cite{Todorov,vanNiekerk,boolean}. In a broader context, \cite{lee2021sunrise} proposes  
to bound ``Bellman updates'', which improves the stability of training and sample efficiency in entropy-regularized RL. However, the method in \cite{lee2021sunrise} does not leverage known solutions for new tasks, instead using a parallel ensemble of learners for variance estimation. 

In the current work we propose a novel method for the derivation of double-sided bounds, which is not limited to a particular type of composition or transformation of prior solution(s), and is valid for an arbitrary function. Our method is a ``zero-shot'' approach for deriving double-sided bounds -- it does not require additional samples beyond those collected by the learning agent. It is applicable to both standard and entropy-regularized RL, to deterministic and stochastic environments, and to discrete and continuous domains. The theoretical results are provided in the following ``Results'' section, and in the ``Applications'' section we demonstrate the applications of the theory in simple domains, leaving large scale experiments to future work.

\section{Results}\label{sec:Theory}
In this section, we focus on entropy-regularized (MaxEnt) RL, the case considered in \cite{Adamczyk_AAAI}. The analogous results for un-regularized RL (which can be considered as a limiting case of entropy-regularized RL) are provided later. The proofs of all results shown can be found in the Appendix.

Our main result provides double-sided bounds on the optimal $Q$ function. We emphasize that \textit{any} (bounded) function $Q: \s \times \A \to \mathbb{R}$ can be used to generate a bound. We suggestively use the notation ``$Q$'' for this otherwise arbitrary function to note that it can be derived from a previous tasks' solution, an estimate, or other ansatz (e.g. composition or hierarchical function) of subtask $Q$-values.

\OnTheFlyBound{}

During training, the Bellman loss $\mathcal{L}=||\Delta||^2 \to 0$, implying that $\inf \Delta \to 0$ and $\sup \Delta \to 0$, hence the bounds in Eq.~\eqref{thm:gen_double_sided_bounds} will become tight upon convergence of the soft action-value function. We note that this is generally \textit{not} the case for un-regularized RL, as will be discussed later.

In principle, given some assumptions on the structure of the reward function or dynamics, it is possible to tighten these bounds. As an example, we provide a tighter lower bound when the MDP always has an ``identity'' action allowing the agent to return to the same state:
\TightOnTheFlyBound{}

In the Appendix, we show that the lower bound of Eq. \eqref{eq:tighter_double_sided} is indeed tighter than Eq. \eqref{eq:gen_double_sided_boundsA} at all state-actions except the minimizer $(s^*,a^*)=\textrm{arginf}\ \Delta(s,a)$.

As an alternative, in practice, one can replace the $\inf$ and $\sup$ in the previous results by a $\min$ and $\max$, respectively, over the finite dataset provided (e.g. the current batch of replay data). Although not exact, this substitution becomes increasingly accurate for large batch sizes. We employ this substitution in the experiments shown in section \ref{sec:experiments}. Nevertheless, we provide an exact extension of our results in the subsequent section for sufficiently well-behaved state-action spaces.

In a similar manner, we may also bound the rate of suboptimality induced by using the policy derived from some estimate $Q(s,a)$:
\SuboptBound{}
This result implies that any policy with a known soft value function has a (lower and upper) bounded suboptimality. The typically-stated objective of minimizing the Bellman loss can be understood as minimizing the suboptimality suffered by the induced policy $\pi \propto \exp \beta Q$.

We conclude this section by showing that a new Bellman operator, which includes clipping when applicable, converges to the optimal $Q$ function:
\ClippedBOperatorThm{}
This result shows that updates with clipping are guaranteed to converge to the same solution. We experimentally demonstrate this in Fig. \ref{fig:maze_learning}.

\subsection{Error Propagation in Continuous Spaces}

The bounds presented in the previous section, though exact, are often intractable due to the required global extremization over continuous state-action spaces. One cannot access the global extrema of $\Delta$ given only finitely many samples in state-action space. Thus, we provide the following bounds, allowing for the extension of our results to (sufficiently well-behaved) continuous spaces. In this section, we loosen those bounds by relaxing the required extremization with a simpler optimization over a given discrete batch of replay data. 

We begin with some helpful definitions.
\begin{definition}
A function $\bar{X}:\s \times \A \to \mathbb{R}$ is an $\varepsilon$-optimal approximation of $X(s,a):\s \times \A \to \mathbb{R}$ if it satisfies $\av{\bar{X}(s,a) - X(s,a)} \leq \varepsilon$ for all $s \in \s, a \in \A$.
\end{definition}
\begin{definition}
The diameter of a bounded metric space, $\mathcal{X}$, endowed with a metric $d(\cdot, \cdot)\to \mathbb{R}_{\geq 0}$ is a constant $D \in \mathbb{R}_{>0}$ such that
$d(x_1, x_2) \leq D$ for all $x_1, x_2 \in \mathcal{X}$.
\end{definition}

\BoundingContinuousExtrema{}

As an example, in the case that one uses the simple upper bound, $Q(s,a) \leq \frac{1}{1-\gamma} \sup r(s,a) $, over a finite-sized batch of replay experience $\{s_i,a_i,r_i, s_{i+1}\}_{i=1}^{T}$, one can bound the (intractable) $\sup$ which is taken over all state-action space: $\sup r(s,a) \leq \min_i r_i + L_r ||(D_\s, D_\A)||_p$. 


In the case of continuous spaces, we cannot calculate the state-value function directly, so one typically resorts to actor-critic methods \cite{Haarnoja_SAC} where a policy network $\pi$ and value network $Q$ are trained together. In this case, one must calculate the entropy-regularized state-value function as $V^\pi(s)=\E_{a \sim{} \pi} \left[ Q^\pi(s,a) - \beta^{-1} \log \pi(a|s) \right]$. However, the expectation over continuously many actions is intractable in the general case. The solution to this is parameterizing the policy network by a simple, but expressive distribution at each state, for instance a Gaussian actor $\mathcal{N}(\mu(s), \sigma(s))$. With knowledge of the means and variances, the sampling error can be bounded as we show below.

\VfuncBound{}
As the policy becomes deterministic ($\sigma \to 0$), in the un-regularized limit ($\beta \sigma \to \infty$), the error reduces to zero as expected (since accurately sampling a deterministic policy only requires one action). Further, the Lipschitz constant in Eq.~\eqref{eq:L_Q_entreg} matches that of the un-regularized case \cite{rachelson2010locality}.
Although the expectation in Eq. \eqref{eq:approx_V} appears intractable, the Gaussian parameterization allows it to be calculable, since the entropy of the policy only depends on its variance.
Under the stated hypotheses, this allows us to translate our bounds in Theorem \ref{thm:gen_double_sided_bounds} to the continuous setting. However, satisfying these hypotheses (e.g. the restriction on $\gamma$) may be challenging in practice. One way of circumventing this is to consider works such as \cite{fazlyab2019efficient}, where one can estimate the Lipschitz constant of the neural net ($Q$-function) being used to generate bounds.

We note that with the Gaussian policy parameterization, the relative entropy (second term in Eq.~\eqref{eq:approx_V}) can be computed exactly from the mean action. In principle, the analysis may be extended to other policy parameterizations. For simplicity, the analysis is carried out for single-dimensional action spaces in the $p=1$ norm, which is easily generalized to other contexts. 

These results allow us to derive the following upper and lower bounds in continuous spaces (an extension of Theorem \ref{thm:gen_double_sided_bounds}), when the $Q$-function used for deriving $\Delta$ is known to be $L_Q$-Lipschitz, or is optimal for an ($L_r, L_p$)-Lipschitz MDP:
\FullyPropagated{}

\subsection{Extension to Un-Regularized RL}\label{sec:std_rl}
Although the previous results have been discussed in the context of entropy-regularized RL, it is possible to extend them to the un-regularized ($\beta \to \infty$) domain as well with the replacement $\Delta' \to r(s,a)+\gamma \E_{s'}V(s') - V(s)$. 
This can be understood as taking the estimated state-value function $V(s)$ to generate a potential function for shaping \cite{ng_shaping} the original reward function $r(s,a)$, with $\Delta'$ now representing this shaped reward. The corresponding value functions are then related by Eq. (3) in \cite{ng_shaping} which can be seen as the analog of Theorem 1 in \cite{Adamczyk_AAAI} for the un-regularized case.
In the Appendix, we show that replacing $\Delta \to \Delta'$ in Theorem \ref{thm:gen_double_sided_bounds}, leaves Eq. \eqref{eq:gen_double_sided_boundsA} and \eqref{eq:gen_double_sided_boundsB} valid for the un-regularized case. In this case, as the Bellman loss decreases, $\mathcal{L} \to 0$, there is no guarantee that $\Delta' \to 0$ as in the regularized case. Interestingly, we nevertheless find that in the un-regularized case, the clipping does occur, and the magnitude of bound violations decreases throughout training.
We use this form (un-regularized RL double-sided clipping) for the FA experiments shown in the next section.

The preceding extension to un-regularized RL can be generalized to address an open problem in research on compositionality. Specifically, we can now address a 
question posed by \cite{Nemecek} concerning the possibility of composing prior solutions in un-regularized RL. We can address this question by deriving an extension of Theorem 10 in \cite{Adamczyk_AAAI} to the case of un-regularized RL.


\StdRLComp{}

Thus, multiple primitive tasks can indeed be composed (via $V_f$) and subsequently corrected (via $K^*$) in un-regularized RL.

\section{Applications}\label{sec:Applications}
The framework developed in this work has applications on both theoretical and experimental fronts. In this section, we discuss some applications relating to compositionality and approaches to clipping. 

\subsection{Exact Composition in Entropy-Regularized RL}
One application of the framework developed is to provide new insights and extensions of previously derived results for value function compositions, as seen in Theorem \ref{thm:std_aaai_}. Previous work \cite{vanNiekerk} on entropy-regularized RL has shown that, for a specific choice of composition function, an exact expression for the optimal value function of interest can be derived. This result can be rederived from a different perspective and also extended to a broader class of compositions using the framework developed. Specifically, we use the composition of value functions for previously solved tasks as an estimate for the optimal value function of the composite task. Then, using this estimate in combination with Theorem \ref{thm:gen_double_sided_bounds}, we derive conditions such that both of the bounds can be saturated with $\Delta(s,a) = 0$, thereby giving an \textit{exact composition}.



Using this approach, we are able to extend the results of \cite{vanNiekerk}, who find an instance of exact composition in entropy-regularized RL for tasks with absorbing states. Our derivation (see Appendix) provides new insight into why specific choices of reward compositions lead to exact compositions of optimal value functions. 
\vanNiekerkExtension{}

A detailed derivation of the result is provided in the Appendix; in the following we note some key points. We consider the setting discussed in \cite{vanNiekerk} (undiscounted, deterministic dynamics with rewards varying only on the absorbing states for the solved tasks). By analyzing the exponentiated version of the backup equation for the solved tasks, we obtain a general class of reward compositions and value function compositions that satisfy the same form of backup equation. The extension from previous work is that the weights no longer need to be normalized to unity.


\begin{figure}[h]
    \centering
    \includegraphics[width=0.5\textwidth]{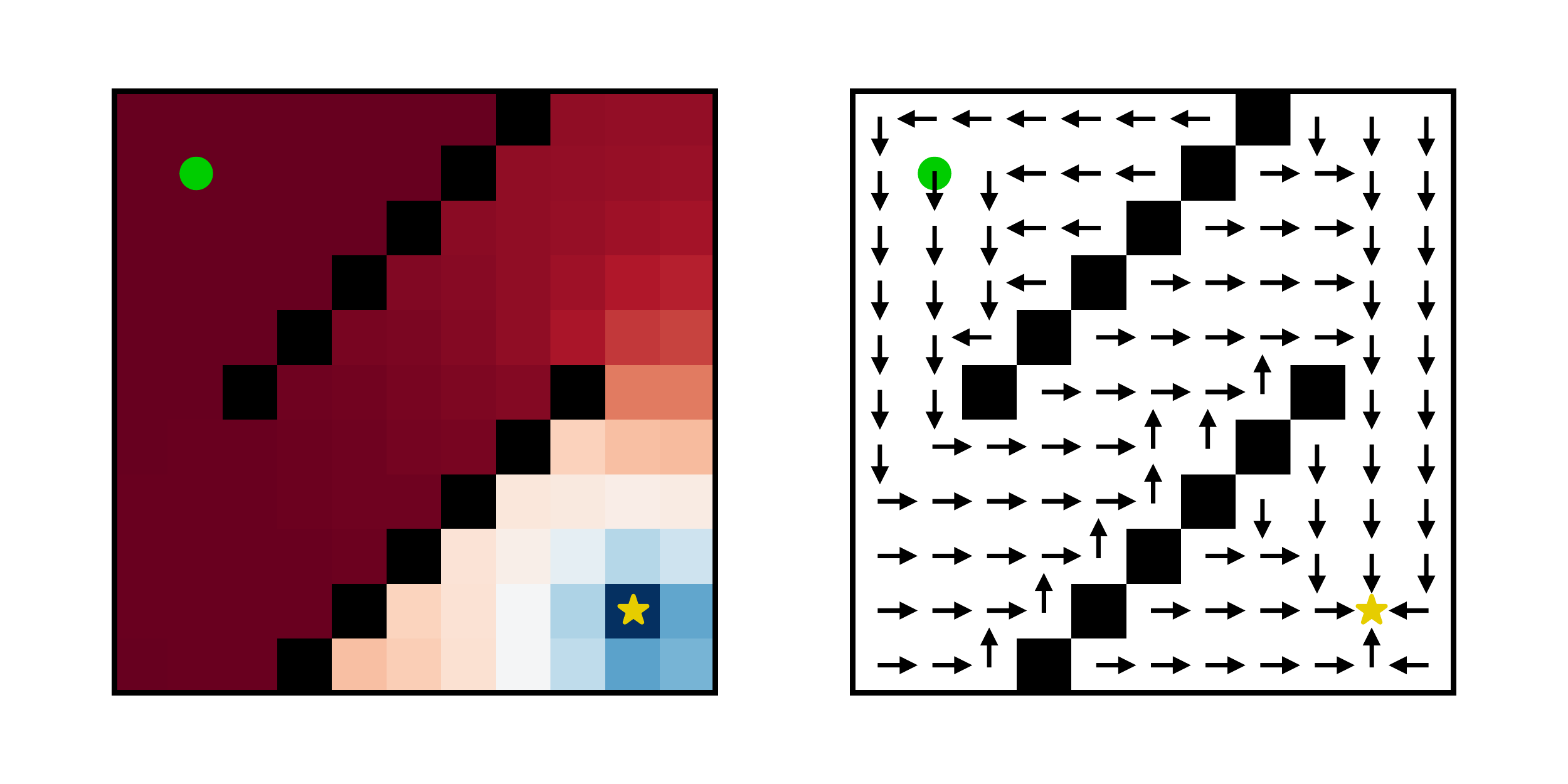}
    \caption{The discrete maze considered for the tabular experiments. The agent begins at the green circle, and the yellow star is the only rewarding state. The action space consists of the cardinal directions, and the state is encoded by the location on the grid. At each step, the agent receives a small penalty if it has not reached the goal. $\gamma=0.98$, $\beta=0.1$. On the left plot, we show the optimal value function $V(s)$ (blue indicates high value). On the right plot, we show the greedy policy extracted from the optimal action value function $\text{argmax}_{a} Q(s,a)$.}
    \label{fig:maze}
\end{figure}

\begin{figure}[ht]
    \centering
    \includegraphics[width=0.45\textwidth]{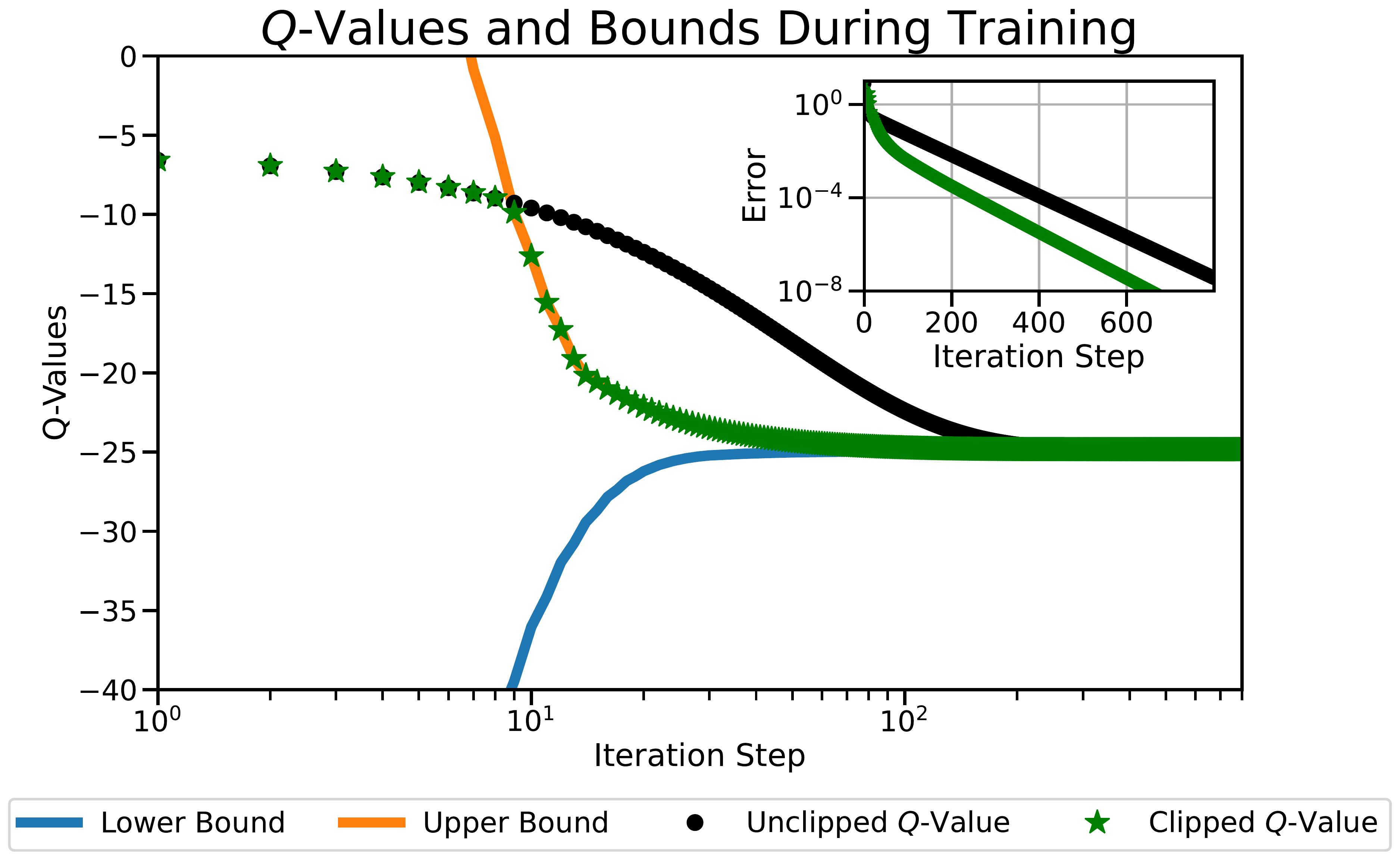}
    \caption{$Q$-values during training with respect to the derived bounds. The error is the maximum difference between consecutive Bellman updates. (Note the $\log$-scaled axes.)}
    \label{fig:maze_learning}
\end{figure}

\subsection{Experimental Validation}\label{sec:experiments}

In the following experiments, we study the utility of clipping based on our theoretical results.
For simplicity, we highlight the results on a simple discrete environment. Without any external estimates for the $Q$ function, we use the estimate given by the previous step's $Q$-function. 

\subsubsection{Tabular Experiments}
In the tabular case, since we have access to the $Q$-table and we perform exact updates, we simply clip the updated $Q$-table according to the derived bounds. In Fig. \ref{fig:maze_learning} we show the results of training in a simple maze environment (Fig. \ref{fig:maze}). In experiments across different sized environments, and with various levels of stochasticity, we universally find the increase in convergence speed shown in the inset plot of Fig. \ref{fig:maze_learning}. In the main plot of Fig. \ref{fig:maze_learning}, we depict the mean $Q$ values over all $(s,a)$ pairs. We find that the violated upper bound (over-optimism) occurs across many tabular domains. In this experiment, we use stochastic transition dynamics with a $50 \%$ probability of taking the intended action and $25\%$ probability of taking an action perpendicular to that intended. As claimed previously, we see that as the Bellman loss reduces (inset plot), the double-sided bounds become tight (blue and orange lines converge).

\subsubsection{Function Approximator Experiments}
In the DQN algorithm used, a target network is employed for stability. We can therefore also use the target network to derive another set of bounds on the true $Q$-values (cf. Appendix for the un-regularized RL bounds corresponding to those given in Theorem \ref{thm:gen_double_sided_bounds}). Since both bounds must hold, we take the tightest bound possible. In general, given many sources of an estimate $Q$-function, one can collectively use them to obtain the tightest bound possible.
\begin{figure}[ht]
    \centering
    \includegraphics[width=0.4\textwidth]{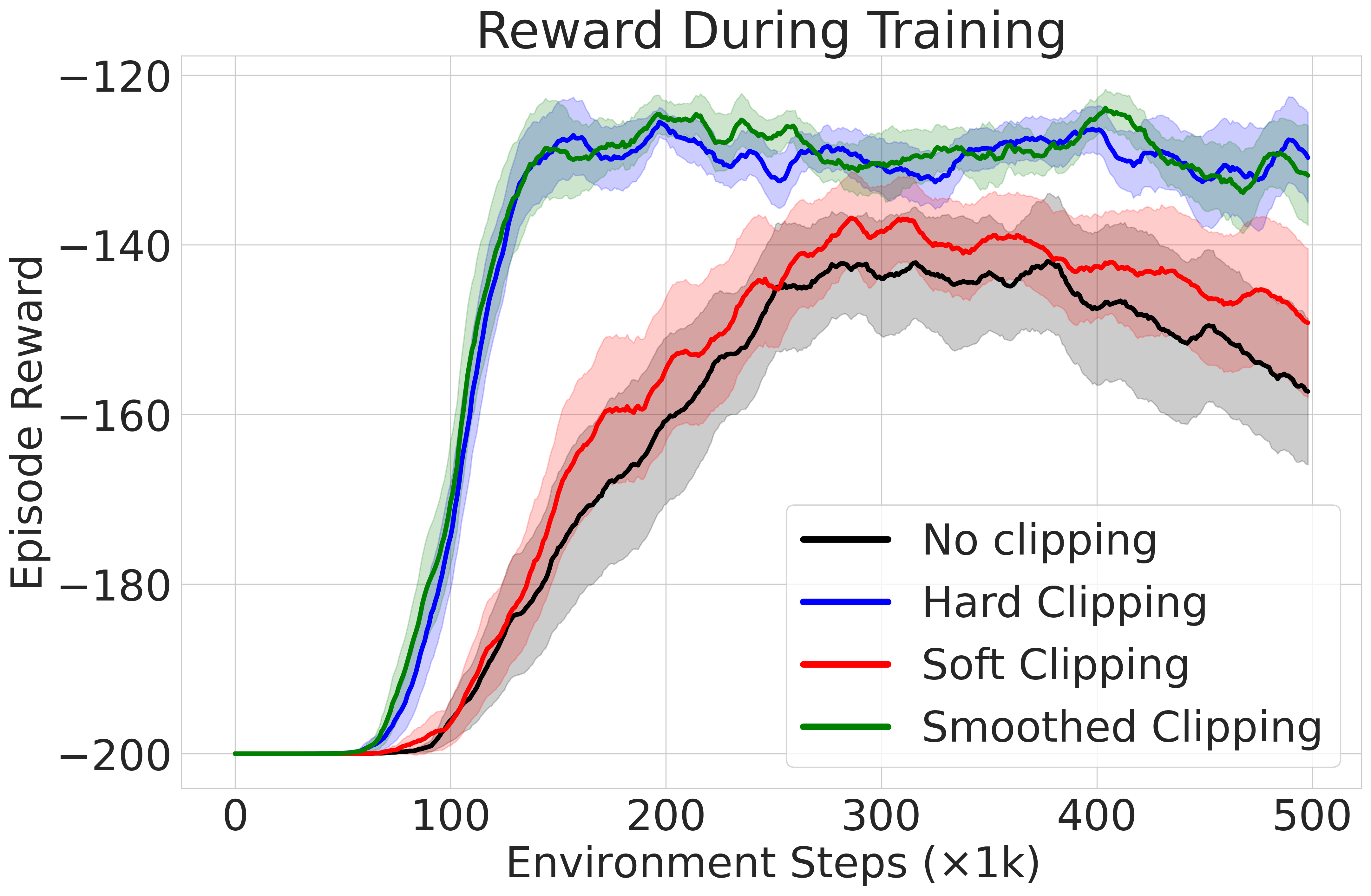}
    \caption{Reward curves for the MountainCar environment. We fine tune each method's hyperparameters, and average over 20 random initializations. The $95 \%$ confidence intervals are shaded for each method.}
    \label{fig:ep_rwds}
\end{figure}
The derived bounds can be implemented using different approaches for clipping of the value function during training. We highlight the different methods used below, inspired by the methods used in \cite{constrainedGPI, Adamczyk_UAI}:
\newline
(0) \textbf{No Clipping:} The standard training scheme for DQN is implemented, with no clipping. 
\newline
(1) \textbf{Hard Clipping:} At each backward pass to the function approximator we enforce the following bounds on the target value:
\begin{equation}
    Q(s,a) \xleftarrow[]{} \hat{Q}(s,a) 
\end{equation}
where $\text{L}$ and $\text{U}$ denote the lower and upper bounds derived in Theorem \ref{thm:gen_double_sided_bounds}, and 
\begin{equation}
    \hat{Q}_{\textrm{clip}} \doteq \min\{ \max\{ r(s,a) + \gamma V(s'),\  \text{L}(s,a) \}, \text{U}(s,a) \}
\label{eq:qclip}
\end{equation}
\newline
(2) \textbf{Soft Clipping:} An additional term, the ``clipping loss'', is added to the function approximator's loss function. The clipping loss is defined as 
\begin{equation}
    \mathcal{L}_{\textrm{clip}}= \av{Q(s,a) - \hat{Q}_{\textrm{clip}}(s,a)}
    \label{eq:clip_loss}
\end{equation}
This gives a total loss of $\mathcal{L}=\mathcal{L}_{\text{Bellman}} + \eta \mathcal{L}_{\text{clip}}$. The hyperparameter $\eta$ weights the relative importance of the bound violations against the Bellman error.  In principle it can be tuned, but we choose to fix $\eta=10^{-5}$ for all experiments, ensuring $\mathcal{L}_{\text{Bellman}} \sim{} \eta \mathcal{L}_{\text{clip}}$. Alternatively, one can view this as equivalent to providing a bonus to the reward function for states with high bound violation. This is analogous to the UCB-style bonus applied in \cite{lee2021sunrise}.
\newline
(3) \textbf{Smoothed Clipping:} The updated $Q$-values are set as an average between those given by Hard Clipping and No Clipping, with a relative weight factor inversely related to the bound violations.
\begin{align*}
    Q(s,a) \xrightarrow[]{} (1-\tau)\left(r(s,a) + \gamma V(s')\right) + 
    \tau \hat{Q}_{\textrm{clip}}(s,a)
\end{align*}
where 
\begin{equation}
    \tau = \frac{\mathcal{L}_{\text{clip}}}{1+\mathcal{L}_{\text{clip}}}
\end{equation}

We note that when the bound violations are zero, the standard update rule is recovered. This value for $\tau$ is chosen to set the relative weight of the two terms to match the magnitude of bound violations: $\tau/(1-\tau)=\mathcal{L}_{\text{clip}}$. Therefore, the clipped values will be preferred over the standard update rule, in direct proportion to the bound violations.

Figure \ref{fig:ep_rwds} indicates that clipping is able to improve the stability and speed of training in the MountainCar environment. Here, we use a bootstrapped estimate of $Q(s,a)$ (that is, the target $Q$-network is bounded by the actively trained $Q$-network).

\section{Discussion}
In summary, we have established a general theoretical framework for deriving double-sided bounds in reinforcement learning. We have explored the use of the double-sided bounds in tabular domains, finding that application of the bounds through clipping is able to speed up training. We also provide some preliminary exploration in the FA domain where new experimental methods for clipping were presented. Furthermore, beyond the theoretical contributions, we believe the current work has the potential to open new directions of research as outlined below.

While the derived bounds are applicable generally to any value function estimate and for arbitrary transition dynamics, it is possible that they are tightened for specific classes of the estimates and restrictions on the dynamics or structure of reward functions. For example, in \cite{Adamczyk_UAI} which analyzed compositions in RL, it was shown that one side of the bound can be simplified further for specific classes of functional transformations or compositions. In future work, it would be interesting to explore under what conditions the bounds may be further simplified or tightened. 

Other promising avenues for future research include: (i) combining our results with ensemble methods such as SUNRISE \cite{lee2021sunrise} which can lead to tighter bounds on the value function, as more estimates are used to derive the double-sided bounds in Theorem \ref{thm:gen_double_sided_bounds}, (ii) using bound violations as a proxy for the best prior task to transfer (minimizing bound violations) when multiple prior solutions are known, (iii) implementing a dynamic schedule for the soft clipping weight parameter, similar to the approach in \cite{Haarnoja_SAC} which includes learning a dynamical temperature parameter.

The extension of \cite{vanNiekerk}'s Theorem 2 (shown above in Theorem \ref{thm:niek}) for value function composition was proved for the case of deterministic dynamics in this work. However, it still remains an open question as to whether this result is generalizable to other domains, e.g. stochastic dynamics. Moreover, other composition methods may yield exact results for the composite task's value function (cf. \cite{boolean, boolean_stoch}). It will be of interest to see if the framework developed in this work can be used to provide insight into the different conditions under which exact compositions can be obtained.

Considering further the composition of multiple previously solved tasks, one can consider the problem of \textit{learning} a composition function $f$, which takes into account the derived bounds. As a learning objective, one could use the magnitude of the difference in bounds, to learn a function $f$ which can be considered an ``optimal composition'' (e.g. related to \cite{PNNs}.

The framework established in this work can be used to obtain bounds for optimal value functions in general settings, not just limited to the composition of tasks.
Specifically, we can use any estimate for the optimal value function as the base knowledge and use the derived results to obtain bounds on the exact optimal value function. In combination with the regret bound derived in this work, iterations of PE/PI can serve as the initial steps in an iterative procedure for progressively improving the bounds to obtain improved approximate solutions. The development of such iterative procedures will be explored in future work. 
}

\newpage
\section{Technical Appendix}
In this technical appendix, we provide further discussion on experimental details and give proofs for all the results shown in the main text. 
\subsection{Experiments} 

In the tabular setting, we perform exact updates of the Bellman backup equation for entropy-regularized RL. At each update step, we calculate the bounds given by Theorem \ref{thm:gen_double_sided_bounds}, which are exact in this case. Then we perform Hard Clipping, by following Eq.~\eqref{eq:qclip} in the main text. Interestingly, we see that as the upper bound becomes tight, the $Q$-values are constantly saturated by this value. The departure of the No Clipping and Hard Clipping $Q$-values is also evident in the reduction of error ($\ell_\infty$ distance) between consecutive iterations.

To explore the utility of clipping in function approximator (FA) systems, we use a DQN learning algorithm \cite{stable-baselines3}, while applying and monitoring clipping given by the bounds in Theorem \ref{thm:gen_double_sided_bounds} for un-regularized RL. In particular, we continuously bootstrap by using the previous estimate of the $Q$-function to generate the bounds, and we clip the target network's output value accordingly. In particular, we extract bounds from both the target network and $Q$-network at each step, and take the tighter of the two bounds. For continuous spaces, we use the estimate $\sup r(s,a) \approx \max_{i \in \mathcal{D}} r(s,a)$, where the $\max$ is taken over the current batch (and similarly for $\inf r(s,a)$). We consider the three clipping methods described in the Experiments section of the main text.

We have also performed the same experiment, with a fixed learning rate, for the Mountain-Car environment \cite{openAI}. These experiments share the hyperparameters shown in Table \ref{tab:shared} and are averaged over 25 runs.

\begin{figure}[h]
    \centering
    \includegraphics[width=0.45\textwidth]{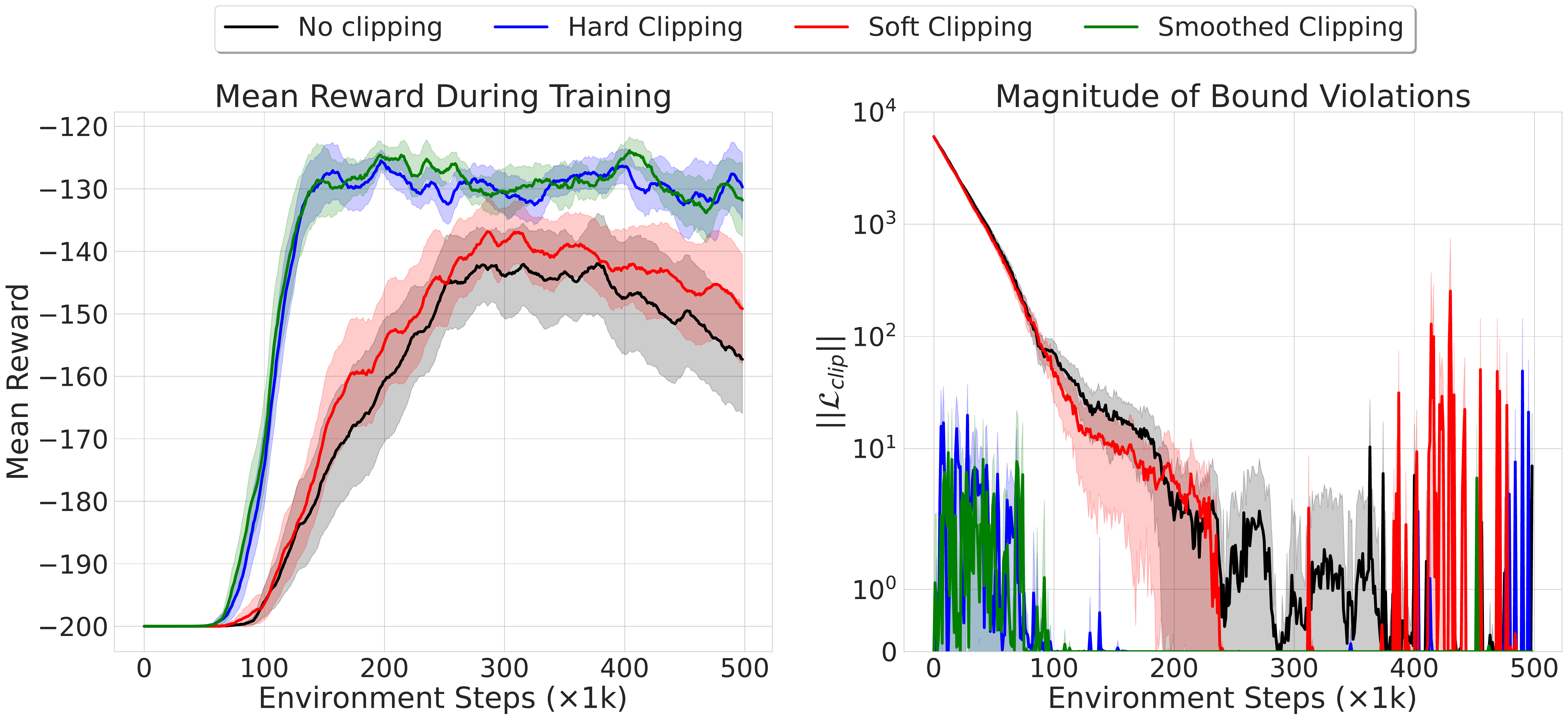}
    \caption{Mountain-Car learning curves for a fixed learning rate $\alpha=0.004$. The mean bound violations and episode rewards throughout training are shown for each clipping method. In the right panel, we plot the total bound violations (magnitude of over- or under-estimation of $Q$ based on the allowed upper and lower bounds). We find that bound violations decrease during training (most quickly for hard and smoothed clipping), which corresponds to better performance in terms of the mean evaluation reward (left plot).}
    \label{fig:mcar_expt}
\end{figure}
We use $\epsilon$-greedy exploration, with a linear schedule from $1.0$ to $0.07$ after $20 \%$ of the total ($N=500 \textrm{k}$) timesteps. The remaining hyperparameters (shared by all clipping methods) are listed below.
\begin{center}

\begin{tabular}{||c c||} 
 \hline
 Hyperparameter & Value \\ [0.5ex] 
 \hline\hline
 Learning Rate & 0.004 \\
 \hline
 Batch Size & 128 \\
 \hline
 Buffer Size & 10,000 \\ 
 \hline
 Discount Factor, $\gamma$ & 0.98 \\
 \hline
 Gradient Steps & 8 \\
  \hline
 Policy Architecture & $(256, 256)$ \\
  \hline
 ``Learning Starts'' & 1,000 \\
 \hline
 Polyak Update, $\tau$ & 1.0 \\
 \hline
 Target Update Interval & 600 \\
 \hline
 Training Frequency & 16 \\
 \hline
\end{tabular}
\label{tab:shared}
\captionof{table}{Hyperparameters shared by all Deep Q Networks. These are the hyperparameters published by the authors of the algorithm used \cite{stable-baselines3}: \url{https://huggingface.co/sb3/dqn-MountainCar-v0}.}
\end{center}

\subsection{Proofs}
In this section we provide proofs of the theoretical results in the main text.
Each proof is prefaced with a restatement of the theorem for the reader's convenience.

We begin with a helpful lemma which bounds the optimal action-value function $Q^*(s,a)$ for any task.
We note that these bounds hold for both un-regularized RL and entropy-regularized RL.
\begin{customlemma}{A}
    For a task with reward function $r(s,a)$, discount factor $\gamma$, the (soft) optimal action-value function $Q^*(s,a)$ satisfies:
    \begin{align*}
        Q^*(s,a) &\geq r(s,a) + \gamma \frac{\inf_{s,a} r(s,a)}{1 - \gamma} \\
        Q^*(s,a) &\leq r(s,a) + \gamma \frac{\sup_{s,a} r(s,a)}{1 - \gamma}
    \end{align*}
    \label{lem:double_sided_bound}
\end{customlemma}

We will prove the upper bound for un-regularized RL, but the proof is identical in entropy-regularized RL and for the lower bound.
\begin{proof}
    The proof follows from induction on the Bellman backup equation:
    \begin{equation}
        Q^{(n+1)}(s,a) = r(s,a) + \gamma\E_{s' \sim{} p(s' | s,a)} \max_{a'} \left( Q^{(n)}(s',a') \right)
    \end{equation}
    The result we aim to prove is the following:
    \begin{align*}
        Q^{(n)}(s,a) &\geq r(s,a) + \gamma \frac{1-\gamma^n}{1 - \gamma}\inf_{s,a} r(s,a)\\
        Q^{(n)}(s,a) &\leq r(s,a) + \gamma \frac{1-\gamma^n}{1 - \gamma}\sup_{s,a} r(s,a)
    \end{align*}
    Since $\lim_{n \to \infty}Q^{(n)}(s,a)=Q^*(s,a)$ and $\gamma \in (0,1)$ the desired result will follow from this limit.

    We set $Q^{(0)}(s,a)=r(s,a)$. The base case ($n=1$) holds as: 
    \begin{align*}
        Q^{(1)}(s,a) &= r(s,a) + \gamma\E_{s' \sim{} p(s' | s,a)} \max_{a'} \left( Q^{(0)}(s',a') \right) \\
        &= r(s,a) + \gamma\E_{s' \sim{} p(s' | s,a)} \max_{a'} r(s',a') \\
        &\leq r(s,a) + \gamma\sup_{s,a} r(s,a) \\
        &= r(s,a) + \gamma \frac{1-\gamma^{1}}{1-\gamma}\sup_{s,a}r(s,a)
    \end{align*}
    We proceed in proving the upper bound.
    For brevity we shall denote $\sup_{s,a} r(s,a) \doteq R$. The inductive hypothesis is
    \begin{equation}
        Q^{(n)}(s,a) \leq r(s,a) + \gamma \frac{1-\gamma^n}{1 - \gamma} R.
    \end{equation}
    To prove that the inequality holds for $n+1$, we use the Bellman backup equation:
    \begin{align*}
        Q^{(n+1)}(s,a) &\leq r(s,a) + \gamma \E_{s'} \max_{a'} \left( r(s',a') + \gamma \frac{1-\gamma^n}{1 - \gamma}R \right)\\
        &\leq r(s,a) + \gamma \left( R + \gamma \frac{1-\gamma^n}{1 - \gamma}R \right)
    \end{align*}
    At this point, if the dynamics model were known then one could improve this bound by including the next term,
    $\E_{s' \sim{} p(s'|s,a)} \max_{a'} r(s',a')$, which we instead bound by $R$.
    Continuing without this term, we have
    \begin{align*}
        Q^{(n+1)}(s,a) &\leq r(s,a) + \gamma \left( R + \gamma \frac{1-\gamma^n}{1 - \gamma}R \right) \\
        &= r(s,a) + \gamma \frac{1-\gamma^{n+1}}{1 - \gamma}R
    \end{align*}
    which completes the proof of the inductive step. As stated above, this completes the proof of the upper bound by taking the limit $n \to \infty$.

    The lower bound follows similarly by swapping all inequalities. The same proof also holds for the soft Bellman backup equation.
\end{proof}

We now proceed with the proof of the first result, Theorem \ref{thm:gen_double_sided_bounds}. We do so by applying Lemma \ref{lem:double_sided_bound} to the $K^*$ function of \cite{Adamczyk_AAAI}'s Theorem 1. 

\OnTheFlyBound{appendix}
\begin{proof}

    As a point of notation, $\widetilde{r}(s,a)$ in \cite{Adamczyk_AAAI} is the same as our $r(s,a)$.
    Using Theorem 1 of \cite{Adamczyk_AAAI}, we have
    \begin{equation}
        Q^*(s,a) = Q(s,a) + K^*(s,a)
    \end{equation}
    where $K^*$ is the optimal soft action value function corresponding to a task with reward function $\Delta(s,a) \doteq r(s,a) + \gamma \E_{s' \sim{}p(\cdot|s,a)} V(s') - Q(s,a)$. 
    By applying Lemma \ref{lem:double_sided_bound} on the value function $K^*$, we arrive at the stated result in Eq.~\eqref{eq:gen_double_sided_boundsBappendix}:
    \begin{align*}
        Q^*(s,a) &= Q(s,a) + K^*(s,a) \\
        &\leq Q(s,a) + \Delta(s,a) + \gamma \frac{\sup \Delta}{1-\gamma} \\
        &= Q(s,a) + r(s,a) + \gamma \E_{s' \sim{}p(\cdot|s,a)} V(s') \\
        &\hspace{4em}- Q(s,a) + \gamma \frac{\sup \Delta}{1-\gamma} \\
        &= r(s,a) + \gamma \left(\E_{s' \sim{}p(\cdot|s,a)} V(s') + \frac{\sup \Delta}{1-\gamma} \right).
    \end{align*}
    A similar proof holds for the lower bound.
\end{proof}

\TightOnTheFlyBound{appendix}
\begin{proof}
    The lower bound in Theorem~\ref{thm:gen_double_sided_bounds} can be tightened by noting that the value function (in both un-regularized and entropy-regularized RL) satisfies a variational form:
    \begin{equation}
        Q(s,a)=\sup_\pi Q^\pi(s,a)
    \end{equation}
    where 
    \begin{equation*}
        Q^\pi(s,a)=\E_{p,\pi} \left[ \sum_{t=0}^\infty \gamma^t r(s_t,a_t) \biggr|\ s_0=s, a_0=a\right]
    \end{equation*}
    and 
    \begin{equation*}
        Q^\pi(s,a)=\E_{p,\pi} \left[ \sum_{t=0}^\infty \gamma^t \left(r(s_t,a_t) - \frac{1}{\beta} \log \frac{\pi(a_t|s_t)}{\pi_0(a_t|s_t)} \right)\right]
    \end{equation*}
    for standard and entropy-regularized RL, respectively (we have dropped the initial state-action conditioning in the latter equation for brevity).
    
    Therefore, one can supply any policy $\pi$ into the objective $Q^\pi$ to obtain a lower bound on the optimal value function. However, the expectation (policy evaluation) is difficult to perform in practice because it corresponds to the solution to another Bellman equation \cite{suttonBook}.
    
    Nevertheless, for particular choices of the input policy $\pi$, one can obtain a simplified expression for $Q^\pi$ leading to a tractable lower bound. With this in mind, we choose the deterministic ``identity policy'', $\pi_\emptyset$, defined as:
    \begin{equation}
        \pi_\emptyset(a|s) = a_\emptyset(s)
    \end{equation}
    where $a_\emptyset(s)$ is the action (for a given state $s\in\s$) such that 
    \begin{equation}
        p(s'|s,a_\emptyset(s)) = \delta(s'-s).
    \end{equation}
    
    In other words, the identity policy is a deterministic policy which transitions the agent back to the same state. We note that this requires the transition dynamics of the task to be deterministic (at least, for this identity action).
    
    With this in mind, we must evaluate the objective $Q^{\pi_\emptyset}~=~\hat{Q}^{\pi_\emptyset}+S^{\pi_\emptyset}$, which we split between the reward and entropic terms. First, we note that since $\pi_\emptyset$ is deterministic, the relative entropy term satisfies
    \begin{equation}
        S^{\pi_\emptyset}=\E_{p,\pi_\emptyset} \left[ \sum_{t=0}^{\infty} \gamma^t \log \frac{\pi_\emptyset(a_t|s_t)}{\pi_0(a_t|s_t)}\right] = 0.
    \end{equation}
    Therefore, it suffices to evaluate the reward contributions alone which can be done as follows:
    \begin{align*}
        \widehat{Q}^{\pi_\emptyset}(s,a) &= \E_{p,\pi_\emptyset} \left[ \sum_{t=0}^\infty \gamma^t r(s_t,a_t) \biggr|\ s_0=s, a_0=a\right] \\
        &= r(s_0,a_0) + \gamma r(s_1, a_\emptyset) + \gamma^2 r(s_1,a_\emptyset) + \dots\\
        &= r(s_0,a_0) + \frac{\gamma}{1-\gamma}r(s_1, a_\emptyset)
    \end{align*}
    We see that the determinism of transitions arising from non-identity actions is required for the first step away from the initial condition. 
    Therefore, we have 
    $Q(s,a) \geq r(s,a) + \frac{\gamma}{1-\gamma}r(s', a_\emptyset)$.
    
    Now, applying this result to the auxiliary task with optimal value function $K^*$:
    \begin{equation}
        K^*(s,a) \geq \Delta(s,a) + \frac{\gamma}{1-\gamma}\Delta(s', a_\emptyset).
    \end{equation}
    Inserting this bound into Theorem 1 of \cite{Adamczyk_AAAI}, we find:
    \begin{align*}
        Q^*(s,a) &\geq Q(s,a) + \Delta(s,a) + \frac{\gamma}{1-\gamma}\Delta(s', a_\emptyset) \\
        &= r(s,a) + \gamma \left( V(s') + \frac{1}{1-\gamma}\Delta(s', a_\emptyset) \right)
    \end{align*}
\end{proof}
As claimed in the main text, we now show that this lower bound is tighter than the previous one in Eq.~ \ref{eq:gen_double_sided_boundsAappendix} of the main text. Since $\Delta(s', a_\emptyset) \geq \inf \Delta(s,a)$, this bound can be saturated only for the initial state-action $(s,a)$ which transitions the agent to $s'=s^*$, the state in which the global reward function $\Delta$ attains its minimum.

\SuboptBound{appendix}
\begin{proof}
    Consider a task with the stated reward function
    \begin{equation*}
    d(s,a)~\doteq~Q^\pi(s,a)~-~\frac{\gamma}{\beta} \E_{s' \sim{} p} \log \E_{a'\sim{} \pi} \exp \beta Q^\pi(s',a').    
    \end{equation*}
    By \cite{cao2021identifiability}, this task's corresponding optimal value function is $Q_d^*(s,a)=Q^\pi(s,a)$. 
    We see that the suboptimality gap $Q^*-Q^\pi$ is nothing but the soft value function $K^*(s,a)$ \cite{Adamczyk_AAAI} for a task with reward function $d(s,a)$, given above. Applying the simple bounds $H \inf d(s,a) \leq K^*(s,a) \leq H\sup d(s,a)$ yields the stated result, with $H=(1-\gamma)^{-1}$ being the time horizon.
\end{proof}

\ClippedBOperatorThm{appendix}

\begin{proof}
    We first show convergence of the operator $\mathcal{B}_C$, then show that it converges to the same fixed point. For convergence, it suffices to show that $|\mathcal{B}_C Q(s,a) - Q^*(s,a)| \leq \gamma |Q(s,a) - Q^*(s,a)|$. 
    
    There are three cases for the magnitude of $\mathcal{B}Q(s,a)$ relative to the upper and lower bounds: 
    \begin{enumerate}
        \item $\mathcal{B}Q(s,a) \in (L(s,a), U(s,a))$
        \item $\mathcal{B}Q(s,a) \in (-\infty, L(s,a))$
        \item $\mathcal{B}Q(s,a) \in (U(s,a), \infty)$
    \end{enumerate}
    
    In the first case, clipping does not occur and hence $\mathcal{B}_CQ(s,a) = \mathcal{B}Q(s,a)$, which contracts with rate $\gamma$.
    In the second case, we can write $\mathcal{B}Q(s,a) = L(s,a) - \chi(s,a)$ where $\chi(s,a) := \mathcal{B}Q - L(s,a) > 0$ is referred to as the ``bound violation''. Then, 
    \begin{align*}
        &\ \ \ \ \ |\mathcal{B}_C Q(s,a) - Q^*(s,a)| \\
        &= |Q^*(s,a)- \mathcal{B}_C Q(s,a)|  \\
        &= |Q^*(s,a) - L(s,a)| \\
        &\leq |Q^*(s,a) - L(s,a) + \chi(s,a)|\\
        &= |Q^*(s,a) - (L(s,a) - \chi(s,a))| \\
        &= |Q^*(s,a) - \mathcal{B}Q(s,a)| \\
        &\leq \gamma |Q(s,a) - Q^*(s,a)|
    \end{align*} A similar proof holds for case 3.
    
    By the Banach fixed point theorem, it follows that repeated application of $\mathcal{B}_C$ converges to a fixed point. It is clear that the fixed point for $\mathcal{B}$ is also a fixed point for $\mathcal{B}_C$, and since it is unique, we have $\mathcal{B}_C^\infty Q(s,a) = \mathcal{B}^\infty Q(s,a) = Q^*(s,a)$.
\end{proof}

\subsection{Error Analysis for Continuous Spaces}
In this subsection, we turn to those results specific to the bounds in continuous spaces and their error analysis, based on Lipschitz-continuity.

\BoundingContinuousExtrema{appendix}
\begin{figure}
    \centering
    \includegraphics[width=0.4\textwidth]{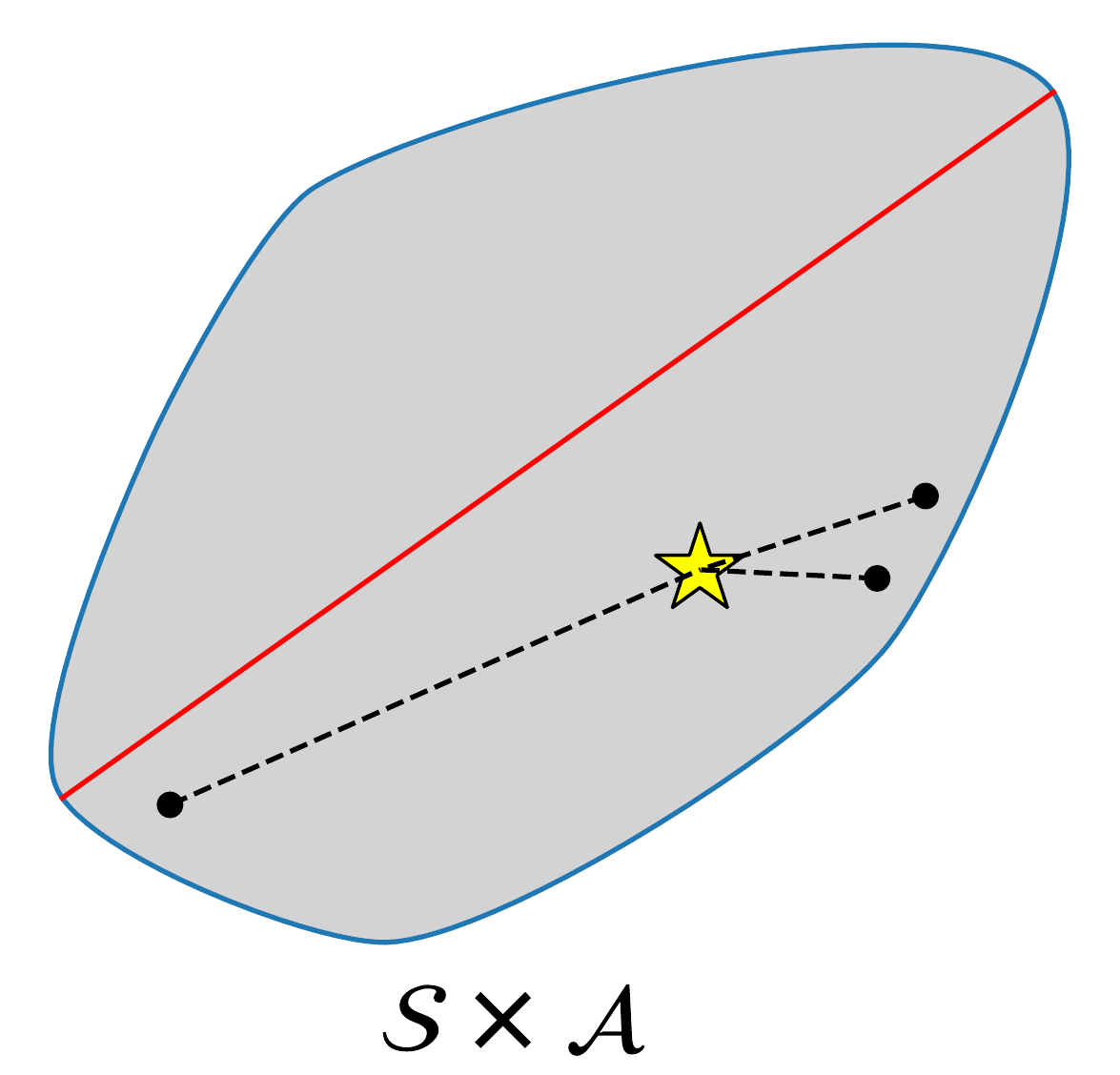}
    \caption{Depiction of a continuous state-action space with a finite set of samples (black points) used to bound the global extrema (star). The diameter of the space is depicted in red. The distance between each sample and the global extrema (dashed lines) is always less than the diameter (solid red line) of the space. Since the growth of the function is linearly bounded by Lipschitz continuity, we can derive a bound on the value of the global extrema given the finitely many samples.}
    \label{fig:extrema_schematic}
\end{figure}

\begin{proof}
    We prove the upper bound on the supremum, the lower bound on the infimum follows similarly. 
    
    Let $\s \times \A$ be a bounded metric space endowed with the $p$-product metric (for simplicity) and let $r:\s \times \A \to \mathbb{R}$ (the function for which we wish to find the global extrema) be $L_r$-Lipschitz continuous. Let the diameters of state and action space be given: $D_\s, D_\A$.
    Suppose a finite set of samples $\mathcal{D} \subset \s \times \A$ is given. Denote $\sup_{s\in \s,a \in \A} r(s,a) =r(s^*, a^*)$. For each $(s,a) \in \mathcal{D}$, the following holds:
    \begin{align*}
        r(s^*,a^*) - r(s,a) &= |r(s^*,a^*) - r(s,a)| \\
        &\leq L_r d\left((s^*,a^*), (s,a) \right)
    \end{align*}
    since the reward function $r$ is $L_r$-Lipschitz in the $d$ metric.
    In practice, the distance between the extrema and an arbitrary point (right-hand side) is unknown, and a generally applicable (albeit loose) bound on this distance is simply the diameter of the space, $D=||(D_{\s}, D_{\A})||_p$. This leads to the following bound:
    \begin{equation}
        r(s^*,a^*)  \leq r(s,a) + L_r D.
    \end{equation}
    This follows from the definition of Lipschitz continuity:
    \begin{align*}
        r(s^*,a^*) - r(s,a) &= |r(s^*,a^*) - r(s,a)|\\
        &\leq L_r d\left((s^*,a^*), (s,a) \right) \\
        &\leq L_r D
    \end{align*}
        
    Since each $(s,a) \in \mathcal{D}$ provides such a bound, we can take the best one (i.e. the minimum over all points in the subset $\mathcal{D}$), recovering the stated bound:
    \begin{equation}
        r(s^*,a^*) \leq \min_{(s,a) \in \mathcal{D}} r(s,a) + L_r D.
    \end{equation}
    In case the calculation $d((s_1,a_1),(s_2,a_2))$ is feasible, one can replace the diameter with the furthest distance from the point in question to any other point in the (bounded) set:
    \begin{equation*}
        r(s^*,a^*) \leq \min_{(s,a) \in \mathcal{D}} \left( r(s,a) + L_r \sup_{s',a'} d((s,a),(s',a'))\right)
    \end{equation*}
    where the $\sup$ is over all $(s,a) \in \s \times \A$.
    which follows by a similar argument as given above:
    \begin{align*}
        r(s^*,a^*) - r(s,a) &= |r(s^*,a^*) - r(s,a)|\\
        &\leq L_r d\left((s^*,a^*), (s,a) \right) \\
        &\leq L_r \sup_{(s',a') \in \s \times \A} d((s,a),(s',a'))
    \end{align*}
This provides a tighter bound but is less tractable in practice.
\end{proof}

We now provide some preliminary results on Lipschitz MDPs which facilitate the proofs of the subsequent results. The following result proves Lipschitz continuity of the value function in un-regularized RL, provided by \cite{rachelson2010locality}.

\begin{customthm}{4.5a}[\citeauthor{rachelson2010locality}]
    Given an $(L_r, L_p)$-Lipschitz continuous MDP and an $L_\pi$-Lipschitz continuous, stationary policy $\pi$, if $\gamma L_p(1+L_\pi) < 1$, then the infinite horizon, $\gamma$-discounted value function $Q^\pi$ is $L_Q$-Lipschitz continuous, with:
    \begin{equation}
        L_Q = \frac{L_r}{1-\gamma L_p (1+L_\pi)}
    \end{equation}
    \label{thm:q_lipschitz}
\end{customthm}
We will extend this result to the case of entropy-regularized RL where the policy's entropy plays a role. To extend it to the entropy-regularized case, we begin with (and following the notation of) Lemma 1 in \cite{rachelson2010locality}. Since the entropy of the policy appears in the calculation of the state-value function, we require a tractable policy class. We use the Gaussian parameterization due to its widespread use \cite{Haarnoja_SAC, stable-baselines3}.
\begin{customlemma}{4.5b}
In entropy-regularized RL, given an $L_Q$-Lipschitz continuous $Q$-function $Q^\pi$ denoting the soft value of a Gaussian policy $\pi(\cdot|s) \sim{} \mathcal{N}\left(\mu(s),\sigma(s)\right)$, the corresponding value
function $V^\pi(s)$ is $L$-Lipschitz continuous, with:
\begin{equation}
    L = L_Q(1+L_\mathcal{N})+\frac{1}{\beta \sigma_{\text{min}}},
\end{equation}
where $\sigma_{\text{min}}=\min_{s}\sigma(s)$ and $L_\mathcal{N} =\sigma_{\text{min}}^{-2}(2\pi e)^{-1/2}$ is the maximum Lipschitz constant of the Gaussian density across all states.
\label{lem:entreg_lipschitz_V}
\end{customlemma}
\begin{proof}
As in SAC \cite{Haarnoja_SAC, stable-baselines3} we assume a Gaussian parameterization with bounded variance $\sigma(s) \geq \sigma_{\textrm{min}}$. We begin by finding the Lipschitz constant for $V^\pi(s)$ in the entropy-regularized setting. Using the definition of the soft state-value function \cite{Haarnoja_SAC},
    \begin{align*}
       &\big|V^\pi(s)-V^\pi(\hat{s})\big| \\
       &\leq \bigg|\E_{a\sim{} \pi }Q^\pi(s,a)- \E_{a\sim{} \pi }Q^\pi(\hat{s},a)\biggr| \\ &+\beta^{-1}\biggl|\left(\mathbb{H}\left[ \pi(\cdot|s) \right] - \mathbb{H}\left[ \pi(\cdot|\hat{s}) \right] \right)\bigg| \\
       &= \bigg|\E_{a\sim{} \pi }Q^\pi(s,a)- \E_{a\sim{} \pi }Q^\pi(\hat{s},a)\biggr| +\beta^{-1}\biggl|\log\frac{\sigma(s)}{\sigma(\hat{s})}\bigg|\\
       &\leq \bigg|\E_{a\sim{} \pi }Q^\pi(s,a)- \E_{a\sim{} \pi }Q^\pi(\hat{s},a)\biggr| +\beta^{-1}\big|\log(s)-\log(\hat{s})\big| \\
       &\leq L_Q(1+L_\pi)\big|s-\hat{s}\big| +\frac{1}{\beta \sigma_\text{min}}\big|s-\hat{s}\big|\\
       &=\left(L_Q(1+L_\pi)+\frac{1}{\beta \sigma_{\text{min}}}\right) \big|s-\hat{s}\big|.\\
    \end{align*}
    The second line follows from the entropy of the Gaussian distribution. The fourth line follows from \cite{rachelson2010locality} and from the Lipschitz-continuity of $\log(\cdot)$ on the domain $(\sigma_{\text{min}}, \infty)$. In practice, one must choose some $\sigma_\text{min}$ to ensure numerical stability. In the case $\beta\sigma_{\min} \to \infty, \sigma_{\text{min}} \to 0$, the policy becomes deterministic and the RL objective reduces to un-regularized RL and the previous result is recovered.

    Since the Gaussian distribution is continuous everywhere, its Lipschitz constant $L_\mathcal{N}=\sigma^{-2}(2\pi e)^{-1/2}$ is easily found by finding the maximum magnitude of the first derivative. Since we are interested in a globally applicable Lipschitz constant, we take the upper bound given by $\sigma_{\text{min}}$. Substituting $L_{\pi} = L_{\mathcal{N}}$ above gives the stated result.
\end{proof}

Now, we extend Lemma 2 of \cite{rachelson2010locality} to the entropy-regularized setting with a Gaussian policy:
\begin{customlemma}{4.5c}
Given an $(L_p, L_r)$-Lipschitz continuous entropy-regularized MDP and a Gaussian policy with bounded variance $\sigma(s) \geq \sigma_{\text{min}}$, the $n$-step, finite horizon, $\gamma$-discounted soft value function $Q^\pi_n$is $L_{Q_{n}}$-Lipschitz continuous and $L_{Q_{n}}$ obeys the recurrence relation
    \begin{equation*}
        L_{Q_{n+1}} = L_r + \gamma \left((1+L_\mathcal{N}) L_{Q_{n}} + (\beta \sigma_{\text{min}})^{-1}\right)L_p
    \end{equation*}
\end{customlemma}

\begin{proof}
The proof is identical to that of Lemma 2 in \cite{rachelson2010locality} except the penultimate line, where we instead use the Lipschitz constant computed for $V^\pi(s)$ in Lemma \ref{lem:entreg_lipschitz_V}:
\begin{align*}
    &\av{Q^\pi_{n+1}(s,a)-Q^\pi_{n+1}(\hat{s},\hat{a})} \\
    &\leq \left(L_r + \gamma L_{V_{n}}L_p \right)\left(|s-\hat{s}|+|a-\hat{a}|\right) \\
    &= \left(L_r + \gamma \left(L_{Q_n}(1+L_\mathcal{N})+\frac{1}{\beta \sigma_{\text{min}}}\right)L_p \right)\times \\
    &\hspace{13em} \left(|s-\hat{s}|+|a-\hat{a}|\right) \\
    &= L_{Q_{n+1}}\left(|s-\hat{s}|+|a-\hat{a}|\right).
\end{align*}

\end{proof}

We are now ready to prove the extension of Theorem \ref{thm:q_lipschitz} in entropy-regularized RL:
\begin{customthm}{4.5d}
    Given an $(L_r, L_p)$-Lipschitz continuous MDP and a Gaussian policy $\mathcal{N}(\mu(s),\sigma(s))$ with bounded variance $\sigma(s) \geq \sigma_{min}$, if $\gamma L_p(1+L_{\mathcal{N}}) < 1$, then the infinite horizon, $\gamma$-discounted value function $Q^\pi$ is $L_Q$-Lipschitz continuous, with:
    \begin{equation}
        L_Q = \frac{L_r + \gamma L_p (\beta \sigma_{\min})^{-1}}{1-\gamma L_p (1+L_\mathcal{N})}
    \end{equation}
    \label{thm:entreg_q_lipschitz}
\end{customthm}
\begin{proof}
We follow the same steps as given in the proof of Theorem 1 of \cite{rachelson2010locality}, concluding by considering the recurrence relation in the convergent limit $L_{Q_{n}}\to L_Q$:
\begin{equation}
    L_{Q} = L_r + \gamma \left((1+L_\mathcal{N}) L_{Q} + (\beta \sigma_{\text{min}})^{-1}\right)L_p
\end{equation}
Solving for $L_Q$ yields 
    \begin{equation}
        L_Q = \frac{L_r + \gamma L_p (\beta \sigma_{\min})^{-1}}{1-\gamma L_p (1+L_\mathcal{N})}.
    \end{equation}
\end{proof}

\VfuncBound{appendix}
\begin{proof}
    We first note that although the relative entropy appears in Eq.~\eqref{eq:approx_V}, we will substitute it with the entropy alone. This is the typical scenario for MaxEnt RL, where the prior policy is ignored. However, in the case of a Gaussian-parameterized prior policy, the remaining term $\E_{a \sim{} \pi}\log\pi_0(a|s)$ has an analytical form. Continuing with the entropy, we see that if the variance is known, it is easily expressed as:
    \begin{equation}
        \mathbb{H}[\mathcal{N}(\mu, \sigma)]=\frac{1}{2}\log(2\pi \sigma^2) + \frac{1}{2}.
    \end{equation}
    Alternative to the variance, the log-probability of the mean is sometimes used in the parameterization \cite{stable-baselines3}, which encodes the same information:
    \begin{equation*}
        -\log(p(\mu)) = -\log\left( \frac{1}{\sqrt{2\pi\sigma^2}}\right)=\mathbb{H}[\mathcal{N}(\mu, \sigma)] - \frac{1}{2}.
    \end{equation*}
    Therefore, we only take into account the error in the first term, the estimation of $\E_{a \sim{} \pi} Q^\pi(s,a)$ given only the mean action $\mu$. We drop the $s$ dependence, denoting $\mu=\mu(s)$ and $\sigma=\sigma(s)$.
    \begin{align*}
        &\ \ \ \ \av{\bar{V}^\pi(s) - V^\pi(s)} \\
        &= \av{\E_{a \sim{} \pi} Q^\pi(s,a) - Q^\pi(s,\mu)} + \av{\E_{a \sim{} \pi} Q^\pi(s,a) - \bar{Q}^\pi(s,\mu)} \\ 
        &\leq \E_{a \sim{} \pi} \left|Q(s,a) - Q(s,\mu)\right| + \varepsilon \\
        &\leq \E_{a \sim{} \pi} L_Q |a - \mu| + \varepsilon\\
        &= \frac{L_Q}{\sqrt{2\pi \sigma^2}}\int_{-\infty}^\infty e^{-\frac{(a-\mu)^2}{2\sigma^2}} |a - \mu| da + \varepsilon\\
        &= \frac{L_Q}{\sqrt{2\pi \sigma^2}} 2\sigma^2 e^{-\mu^2/2\sigma^2} + \varepsilon \\
        &= \sqrt{\frac{2}{\pi}}L_Q \sigma e^{-\mu^2/2\sigma^2} + \varepsilon
    \end{align*}
    Here we have used the one-dimensional absolute value norm for actions, but the result can be readily extended in a similar way for particular choices of the metric on the action space.
    The fifth line follows from the $Q$ function being $L_Q$-Lipschitz continuous, and the final line follows from substituting in Theorem \ref{thm:q_lipschitz} for $L_Q$.
\end{proof}

Interestingly, this result has shown that there is maximum potential error obtained in iterations of policy evaluation, with a non-trivial dependence on the variance of the distribution in question.

To prove Theorem \ref{thm:fully_propagated} we first provide some lemmas detailing the error analysis for the $V^\pi(s)$ and $\Delta(s,a)$ terms appearing in the double-sided bounds of Theorem \ref{thm:gen_double_sided_bounds} and Lemma \ref{lem:extrema}; both of which are prone to estimation errors. 

\begin{customlemma}{4.6a}
    The maximum error in replacing $\Delta$ with $\bar{\Delta}$ (as defined in Theorem \ref{thm:fully_propagated}, i.e. by using the one-point estimate for the expected $Q$-value) is upper bounded:
    \begin{equation*}
        |\Delta(s,a)-\bar{\Delta}(s,a)| \leq\gamma \sqrt{\frac{2}{\pi}} L_Q \E_{s'\sim{} p} A(s') 
    \end{equation*}
    where we introduce the shorthand $A(s)=\sigma(s) e^{-\mu(s)^2/2\sigma(s)^2} + \varepsilon$.
    \label{lem:delta_err_bound}
\end{customlemma}

\begin{proof}
\begin{align*}
    &\ \ \ \ |\Delta(s,a)-\bar{\Delta}(s,a)| \\
    &= \gamma \big| \E_{s' \sim{} p} \left(V(s') - \bar{V}(s')\right) \big| \\
    &\leq \gamma \E_{s' \sim{} p} \big| V(s') - \bar{V}(s')\big| \\
    &\leq \gamma \left(\sqrt{\frac{2}{\pi}} L_Q \E_{s' \sim{} p} \sigma(s') e^{-\mu(s')^2/2\sigma(s')^2} +\varepsilon\right)
\end{align*}
\end{proof}

\begin{customlemma}{4.6b}
The reward function $\Delta$ generated from an $L_Q$-Lipschitz continuous function $Q(s,a)$, $\Delta(s,a)~\doteq~r(s,a)~+~\gamma~\E_{s'} V(s') - Q(s,a)$ with ($L_r, L_p$)-Lipschitz rewards and dynamics, is Lipschitz continuous with
\begin{equation*}
    L_\Delta=\max \left\{L_r, L_Q, \gamma L_p \left(L_Q(1+L_\mathcal{N}) + (\beta \sigma_{\text{min}})^{-1}\right)\right\}.
\end{equation*}
\end{customlemma}
\begin{proof}
The Lipschitz constant of a sum of Lipschitz functions: $r(s,a)+\gamma \E_{s'} \bar{V}(s') - Q(s,a)$ is itself Lipschitz continuous, with the Lipschitz constant being the maximum of all terms' Lipschitz constants:
\begin{equation}
    L_\Delta=\max \left\{L_r, L_Q, \gamma L_p L_V \right\},
\end{equation}
where $L_V$ is given in Lemma \ref{lem:entreg_lipschitz_V}.
Since the relative magnitude of each Lipschitz constant is unknown a prior, we can make no further simplification without additional assumptions.
\end{proof}

Now we are positioned to prove Theorem \ref{thm:fully_propagated}, the double-sided bounds on the soft $Q$-function with estimation errors included.
\FullyPropagated{appendix}

\begin{proof}
We will prove the upper bound, with the lower bound following accordingly.

Beginning with the exact form in Theorem \ref{thm:gen_double_sided_bounds}, the main idea is to propagate the errors due to the single-point estimation for $\bar{V}$, the resulting error in the calculation of $\Delta$ itself, and the $\sup(\Delta)$ estimation.
\begin{align*}
    Q(s,a) &\leq r(s,a)+\gamma \left(\E_{s' \sim{} p} V^\pi(s') + \frac{\sup \Delta(s,a)}{1-\gamma} \right) \\
    &\leq r(s,a)+\gamma \E_{s' \sim{} p}\biggr[\big| V^\pi(s') -\bar{V}^\pi(s')\big| + \bar{V}^\pi(s') \biggr] \\
    &\hspace{4em}+ \frac{\gamma}{1-\gamma}\left(\min_{(s,a) \in \mathcal{D}} \Delta(s,a) + L_\Delta D \right)  \\
    &\leq r(s,a) + \gamma \E_{s'\sim{} p}\left[ \bar{V}^\pi(s') + A(s')\right] \\
    & \hspace{4em}+  \frac{\gamma}{1-\gamma}\left(\min_{(s,a) \in \mathcal{D}} \Delta(s,a) + L_\Delta D  \right) \\
    &\leq r(s,a) + \gamma \E_{s' \sim{} p}\left[\bar{V}^\pi(s') + A(s')\right] \\     
    &\hspace{-2em}+\frac{\gamma}{1-\gamma}\left(\min_{(s,a) \in \mathcal{D}} \left(\bar{\Delta}(s,a) + \gamma \E_{s' \sim{} p} A(s') \right) + L_\Delta D  \right) \\
\end{align*}
where $A(s)=\sqrt{\frac{2}{\pi}}L_Q \sigma(s) e^{-\mu(s)^2/2\sigma(s)^2}+\varepsilon$ and $L_\Delta=\max \left\{L_r, L_Q, \gamma L_p L_V\right\}$. 
The second line follows from Lemma \ref{lem:extrema}, the third line follows from Theorem \ref{thm:vfunc_bound}, and the fourth line follows from Lemma \ref{lem:delta_err_bound}.
\end{proof}

\subsection*{Un-Regularized RL}
We now turn to proofs of the analogous results in standard (un-regularized) RL. We begin by using \cite{ng_shaping} to connect to the results of \cite{Adamczyk_AAAI} and \cite{cao2021identifiability}.
In un-regularized RL \cite{Adamczyk_AAAI} Theorem 1 holds,

\StdRLRwdChange{appendix}
\begin{proof}
    Since $\kappa(s,a)$ is simply the reward function $\widetilde{r}(s,a)$ shaped by the potential function $V^*(s)$, this is simply a re-writing of Eq.~(3) in \cite{ng_shaping}.
\end{proof}

Now we provide a lemma before proving a similar result for compositions. Motivated by \cite{cao2021identifiability}'s~\CaoThmNospace, we provide the same result for standard (un-regularized) RL:
\begin{customlemma}{4.8a}
    Let $Q(s,a)$ be given. Define $V^*(s) = \max_{a}Q(s,a)$ as the corresponding state value functions for a un-regularized RL task.
    Then 
    \begin{equation}
        R(s,a) = Q(s,a) - \gamma \E_{s' \sim{} p} V^*(s')
    \end{equation}
    \label{lem:stdRL_rwdchange}
    is the reward function for a task with optimal action-value function $Q^*(s,a)=Q(s,a)$.
\end{customlemma}
\begin{proof}
    The proof is trivial, given by rearrangement of the Bellman optimality equation.
\end{proof}

\StdRLComp{appendix}
\begin{proof}
    Let $\fQ$ stand for the primitive task's solution, as in Theorem \ref{thm:rwd_change_stdRLappendix}.
    Then, by Lemma \ref{lem:stdRL_rwdchange}, such a value function is optimal for a un-regularized RL task
    with reward function $R(s,a) = \fQ - \gamma \E_{s' \sim{} p} V_f^*(s')$, where $V_f(s) = \max_{a} \fQ$.
    By Theorem \ref{thm:rwd_change_stdRLappendix}, the corrective task has a reward function
    \begin{equation}
        \kappa(s,a) = \fr + \gamma \E_{s'} V_f(s') - V_f(s)
    \end{equation}
    with corresponding optimal value function $K^*(s,a)$, related to $\widetilde{Q}^*(s,a)$ by
    \begin{equation}
        \widetilde{Q}^*(s,a) = V_f(s) + K^*(s,a)
    \end{equation}
    Again, this result can be seen as \cite{ng_shaping}'s reward shaping with a potential function $\Phi(s)=V_f(s)$.
\end{proof}

We now note that Lemma \ref{lem:double_sided_bound} applies to the cases of Theorem \ref{thm:rwd_change_stdRLappendix} and \ref{thm:std_aaai_appendix}, which results in double-sided bounds given \textit{any} estimate of the \textit{state value function} $V(s)$:

\OnTheFlyBoundStd{appendix}
\begin{proof}
The proof is identical to that of Theorem \ref{thm:gen_double_sided_bounds}, except with the proper replacement of $\Delta$.
\end{proof}

\section{Exact composition in entropy regularized RL }

Here, we provide a new proof and extension of Theorem 2 in \cite{vanNiekerk} to highlight that our results can provide new insight to \textit{exact} compositions in entropy-regularized RL.


To align with the assumptions of \cite{vanNiekerk}, we consider the undiscounted, finite horizon setting with deterministic dynamics. We first note the observation which forms the starting point of our analysis: the difference between the true optimal value function ($Q^*(s,a)$), corresponding to reward function $r(s,a)$, and any estimate of the value function ($Q(s,a)$) can itself be represented as another optimal value function, with the corresponding reward function given by~\cite{Adamczyk_AAAI}:
\begin{equation*}
    \Delta(s,a) \doteq r(s,a) + \gamma V(s') - Q(s,a)
\end{equation*}
It is straightforward to show that this observation remains valid in the undiscounted ($\gamma =1$) setting as well. Now, if the estimate of the value function is exact, we must have $\Delta(s,a)=0$. In the following, we determine conditions which lead to $\Delta(s,a)=0$ and correspondingly to exact compositions.

\begin{proof}
We consider $M$ solved tasks with reward functions $\{r_1,\dotsc,r_M\}$ varying only on the set of absorbing states ($s \in \G$). Let $Q_{i}(s,a)$ denote the optimal value function for the $i^{\mathrm{th}}$ task. Consider the composite task with the following reward structure:
\begin{itemize}
 \item For the absorbing states ($s \in \G$), the reward function is given by the reward composition function $\widetilde{r}(s,a) = g(\{r_i(s,a)\})$. 
\item  For the interior states ($s \not\in \G$), the reward function is taken to be the same as the solved tasks and will be denoted by $r(s,a)$.
 \end{itemize}
For the composite task defined in this way, we wish to determine if the corresponding optimal value function can be expressed exactly as some global composition of the known value functions for the solved tasks, denoted by $f(\{Q_i(s,a)\})$. In other words, the estimate of the optimal value function is given by $f(\{Q_i(s,a)\})$, and we will show how a specific form for $f$ corresponds to $\Delta(s,a)=0$ (exact composition).

In the following, we will first show that we must have $f=g$, i.e the value composition function 
must be identical to the reward composition function for the absorbing states.
We will then determine a specific form of $f(\{Q(s,a)\})$ such that the corresponding reward function (i.e. $f(\{Q(s,a)\}) - V_f(s')$, by \cite{cao2021identifiability}) is equal to the reward function for the composite task ($\widetilde{r}(s,a)$), thus yielding $\Delta(s,a) = 0$.
We will do so by deriving the soft back-up equation for $f(\{Q(s,a)\})$ using the soft back-up equations for the subtasks. \\

We begin by observing that, on the absorbing set $\G$, we have $r(s,a)=Q(s,a)$ for all $s \in \G$, implying that $\widetilde{Q}(s,a)=\widetilde{r}(s,a)=g(\{r_i(s,a)\})=g(\{Q_i(s,a)\})$. Thus, for exact composition on the absorbing set $\G$, the value composition function must be the same as the reward composition function (i.e $f=g$), for any reward composition function $g$. Since we are interested in a global value composition function, this means that the reward composition function $g$ also determines the composition function $f(\{Q(s,a)\})$ for states $s \not\in \G$. However, for arbitrary choices of $g$, the corresponding $f(\{Q(s,a)\})$ will not, in general, correspond to the exact optimal value function for states $s \not\in \G$.

We now consider a special class of reward composition functions $g$, such that the corresponding value composition function $f$ is an exact composition globally. Consider $g$ such that we have, for the absorbing states $s$, 
\begin{equation}
        e^{\widetilde{r}(s,a)} =  \sum_i w_i e^{r_i(s,a)} 
\end{equation}
with weights $w_{i} > 0$ and we have set $\tau=1$ for simplicity.

For deterministic dynamics, focusing on the non-absorbing states (i.e. $s \not\in \G$ ) the soft backup equation for the subtask $m$ can be expressed as 
\begin{equation}
    e^{Q_m(s,a)} = e^{r_m(s,a)} e^{V_m(s')}.
\end{equation}
Since the subtask reward functions are identical for those $s \not\in \G$, this simplifies to
\begin{equation}
    e^{Q_m(s,a)} = e^{r(s,a)} e^{V_m(s')}.
\end{equation}

Since the state space is made of disjoint absorbing and non-absorbing (i.e. boundary and interior as in~\cite{Todorov}), we can split to two cases as $s,a$ which transition to $s' \in \G$ and otherwise.






Now, consider the backup equation for each subtask, where we split those states $s \not\in \G$ and $s \in \G$.
\begin{align*}
    e^{Q_i(s,a)} &= e^{r(s,a)} \times \\
    &\left( \sum_{s' \in \G} p(s'|s,a) e^{V_i(s')} + \sum_{s' \not\in \G} p(s'|s,a) e^{V_i(s')} \right)
\end{align*}

But for $s \in \G$, the state value function is simply $V_m(s)=r_m(s)$.
Thus we have 
\begin{align}
    e^{Q_i(s,a)} &= e^{r(s,a)}\times  \label{eq:Qibackup}\\
    &\left( \sum_{s' \in \G} p(s'|s,a) e^{r_i(s')} + \sum_{s' \not\in \G} p(s'|s,a) e^{V_i(s')} \right)  \nonumber
\end{align}


Now, since we have $f=g$, the optimal value composition function is given by
\begin{equation}
 e^{f(\{Q_i(s,a)\})} = \sum_i w_i e^{Q_i(s,a)}  \label{eq:f(Qi)}
\end{equation}
Multiplying each of the subtask backup equations (above) by the respective weight ($w_i$) and summing up we obtain 
\begin{align*}
    &e^{f(\{Q_i(s,a)\})} = e^{r(s,a)} \times \\
    &\sum_i w_i \biggl( \sum_{s' \in \G} p(s'|s,a) e^{r_i(s')} + \sum_{s' \not\in \G} p(s'|s,a) e^{V_i(s')} \biggr). \\
\end{align*}

Now we observe that for $f$ as defined above, the soft state-value function $V_f(s)$ derived from $f(Q)$ satisfies:
\begin{align*}
    e^{V_f(s)} &= \E_{a'\sim{} \pi_0} ~e^{f(\{Q_i(s',a')\})} \\
               &= \E_{a'\sim{} \pi_0} \sum_i w_i  ~e^{Q_i(s',a')} \\
               &= \sum_i w_i ~\E_{a'\sim{} \pi_0} ~e^{Q_i(s',a')} \\
               &= \sum_i w_i ~e^{V_i(s')}
\end{align*}
Using the above, we obtain
\begin{align*}
    &e^{f(\{Q_i(s,a)\})} = e^{r(s,a)} \times \\
    &\left( \sum_{s' \in \G} p(s'|s,a) e^{\widetilde{r}(s,a)} + \sum_{s' \not\in \G} p(s'|s,a) e^{V_f(s')} \right)
\end{align*}


Comparing the above equation with the backup equation for the subtask Eq.~\eqref{eq:Qibackup}, we obtain that $f(\{Q_i(s,a)\})$ (defined in Eq.~\eqref{eq:f(Qi)}) is the exact optimal value function for the composite task with reward function $\widetilde{r}(s,a)$ for the absorbing states ($s \in \G$) and $r(s,a)$ for the non-absorbing states ($s \not \in \G$). The result stated in the main text (Theorem 5.1) follows, given that $\widetilde{Q}(s,a) = f(\{Q_i(s,a)\})$.
\end{proof}

\nocite{openAI}
\nocite{lee2021sunrise}

\end{document}